\documentclass{article}

\usepackage[preprint]{neurips_2024}
\usepackage{natbib}
\setcitestyle{numbers,square}

\usepackage[utf8]{inputenc} 
\usepackage[T1]{fontenc}    
\usepackage{hyperref}       
\usepackage{url}            
\usepackage{booktabs}       
\usepackage{amsfonts}       
\usepackage{nicefrac}       
\usepackage{microtype}      
\usepackage{xcolor}         
\usepackage{svg}
\usepackage{amsmath}
\usepackage{multirow}
\usepackage{graphicx}
\usepackage{subcaption}
\usepackage{threeparttable}
\usepackage{float}

\title{Long Term Memory : The Foundation of AI Self-Evolution }

\author{
\textbf{Xun JIANG}$^{\mu}$$^{\theta}$ \qquad
\textbf{Feng LI}$^{\theta}$\thanks{Equal contribution.} \qquad
\textbf{Han ZHAO}$^{\theta}$\footnotemark[1] \qquad
\textbf{Jiahao QIU}$^{\iota}$\footnotemark[1] \qquad
\textbf{Jiaying WANG}$^{\theta}$\footnotemark[1] \qquad
\\
\textbf{Jun SHAO}$^{\theta}$\footnotemark[1] \qquad 
\textbf{Shihao XU}$^{\theta}$\footnotemark[1] \qquad
\textbf{Shu ZHANG}$^{\theta}$\footnotemark[1] \qquad
\textbf{Weiling CHEN}$^{\theta}$\footnotemark[1] \qquad
\textbf{Xavier TANG}$^{\theta}$\footnotemark[1] \qquad
\\
\textbf{Yize CHEN}$^{\theta}$\footnotemark[1] \qquad
\textbf{Mengyue WU}$^{\alpha}$ \qquad
\textbf{Weizhi MA}$^{\sigma}$ \qquad
\textbf{Mengdi WANG}$^{\iota}$ \qquad
\textbf{Tianqiao CHEN}$^{\mu}$$^{\theta}$ \\
\\
$^\mu$ Tianqiao and Chrissy Chen Institute \\
$^\iota$ Princeton University \\
$^\sigma$ Institute for AI Industry Research, Tsinghua University \\
$^\alpha$ Shanghai Jiao Tong University \\
$^\theta$ Shanda Group \\
}

\begin{document}

\maketitle
\begin{abstract}
Large language models (LLMs) like GPTs, built on vast datasets, have demonstrated impressive capabilities in language understanding, reasoning, and planning, achieving performance comparable to humans in various challenging tasks. Most studies have focused on further enhancing these models by training them on ever-larger datasets, aiming to develop more powerful foundation models. However, while training stronger foundation models is crucial, we propose how to enable models to evolve while inference is also vital for the development of AI, which refers to AI \textit{self-evolution}.
%
Compared to using large-scale data to train the models, the self-evolution may only use limited data or interactions. Drawing inspiration from the columnar organization of the human cerebral cortex, we hypothesize that AI models could potentially develop emergent cognitive capabilities and construct internal representational models through iterative interactions with their environment. 
%
To achieve this, we propose that models must be equipped with Long-Term Memory (LTM), which stores and manages processed real-world interaction data. LTM not only enables the representation of long-tail individual data in statistical models but also facilitates self-evolution by supporting diverse experiences across various environments and agents.
%
In this report, we first explore the concept and significance of AI Self-Evolution, focusing on its potential to enhance AI models during the inference stage. We examine the role of LTM as a key mechanism for enabling lifelong learning in AI systems, allowing models to continually evolve based on accumulated interactions and experiences. Next, we detail the structure of LTM and the corresponding data systems required to facilitate high-quality data acquisition and retention, ensuring the effective representation of individual data. Finally, we classify various approaches for constructing personalized models using LTM data and discuss how models enhanced by LTM can achieve self-evolution through interaction with their environments.Based on LTM, our multi-agent framework OMNE achieved first place on the GAIA benchmark. This demonstrates the great potential of utilizing LTM for AI Self-Evolution and solving real-world problems.
%
We present our technical roadmap and discuss potential avenues for future research. We believe that advancing research in LTM is critical for the ongoing development and practical application of AI technology, especially for self-evolution. We hope this work will inspire more researchers to contribute to the exploration of this exciting and evolving field.
\end{abstract}

\section{Introduction}
%


\begin{figure}
    \centering
    \includegraphics[width=0.8\textwidth]{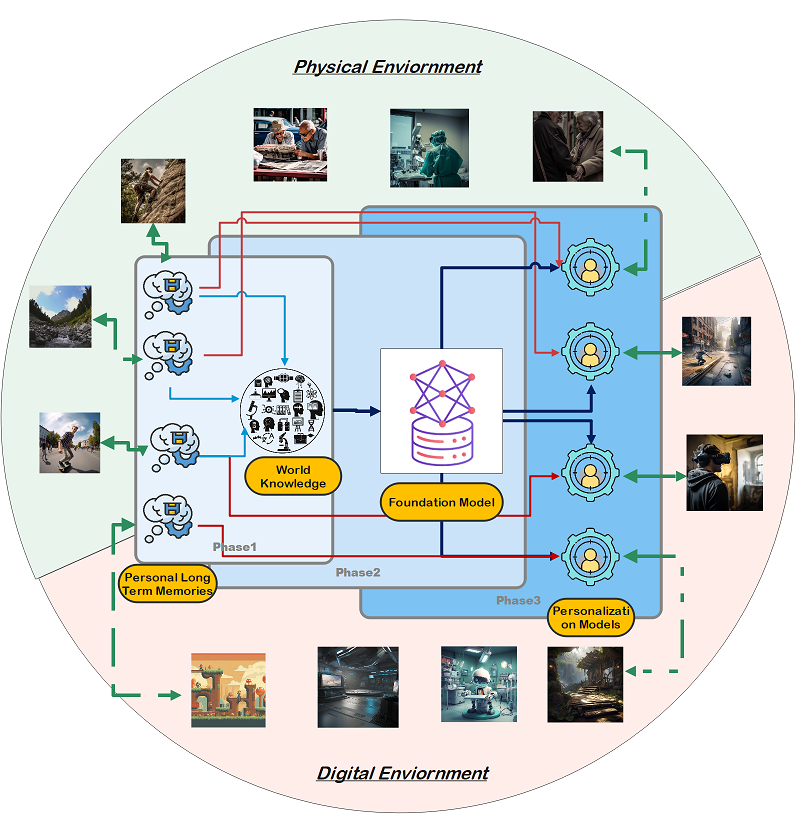}
    \caption{The overview of LTM and AI Self-Evolution.}
    \label{fig:sec1_overview}
\end{figure}

Artificial Intelligence (AI) is recognized as a key technology in the Fourth Industrial Revolution\cite{XuDavidKim2018}, empowering machines to perceive their environments and act intelligently through algorithms and software to optimize the achievement of various objectives. AI technologies are now widely applied in areas such as finance, education, and healthcare. In recent years, the development of large language models (LLMs) and LLM-powered agents has significantly enhanced AI capabilities, which are more powerful in solving various challenging tasks under diverse scenarios.

At its core, a model can be understood as an advanced form of data compression. A classic example is Newton's law of universal gravitation, which condenses complex astronomical data into a simple mathematical formula. This compression represents large amounts of data in a concise form. Similarly, LLMs compress vast amounts of text corpora into statistical patterns to generate coherent text~\cite{doddapaneni2024userembeddingmodelpersonalized}. However, we argue that intelligence is not limited to learning from existing data, the self-evolution ability is also important for the development of AI models, which is similar to the evolution ability of humans. Tasks from different scenarios often have distinct data distributions and diverse ability requirements, where the self-evolution ability will enable AI models to adapt to new tasks by learning limited interactions for powerful performances. Self-evolution will also contribute to diverse models, which can further help the development of AI, especially LLM models in recent years. 


\subsection{Phases of Model Evolution processes of LLMs}
\label{chap:avg_to_personalization_llm}

To better understand the need for self-evolution for LLMs, as shown in Figure \ref{fig:sec1_overview}, we propose to break down the model evolutionary process of LLMs into three main phases. These phases highlight the gradual progression from simple pattern recognition to self-evolved personalized intelligence.
\begin{itemize} \item 
    \textbf{Phase 1: Cognitive Accumulation in the Physical World.} Data accumulation is the first and vital step for the development of AI, which is achieved by humans through continuous practical interactions. In the process of understanding the world, humans have also evolved stronger abilities to discover and apply patterns, but the first step is cognitive accumulation. Individuals interact with their environment, producing diverse and personal fragmented cognition pieces. Some of these cognitive fragments are digitized and stored, which can be used to construct AI model. While others remain in individual minds, contributing to diverse personalities. 
    \item \textbf{Phase 2: Constructing Foundation Models in the Digital World.} AI models attempt to learn from the data accumulated by humans in Phase 1, and have achieved promising results, especially LLMs. LLMs consolidate all digitized cognitive fragments to form a unified ``average'' model (foundation model). These models reflect commonalities and general patterns in large-scale data, making them suitable for a broad range of language generation tasks. However, these models, while statistically efficient, overlook the expression of personalized information and struggle with handling long-tail data or rare scenarios. We think the main reason is they ignored the remained individual evolution of humans.
    \item \textbf{Phase 3: Model Self-Evolution to Achieve Stronger Intelligence.} The third phase moves beyond averaged intelligence, focusing on building self-evolving, personalized intelligent models. To address the complexity and sparsity of personalized data, future model architectures must break away from the existing ``global average'' paradigm and shift towards more flexible and adaptive distributed intelligence architectures, even with limited interactions in a new task/scenario. This self-evolution ability will also contribute to more diverse and stronger intelligence models through their dynamic and continual evolution. Furthermore, the most promising thing is multi-agent evolution based on single self-evolution.
\end{itemize}

Most existing work focuses on how to construct better data and use it to train a more powerful foundational model, which is essentially research centered around phases 1 and 2. There is also currently a popular view that \textit{Architectures aren't fundamentally important in the curve-fitting paradigm, while the critical factor is the dataset}~\cite{Brisbin1987}. This perspective applies to the second phase, but in the third phase, architecture becomes as important as data. The core challenge lies in \textbf{how to effectively express small amounts of individual data within the foundation of statistical models.} Our research focuses on how to distill individual data to ensure more efficient expression within statistical models. At the same time, we are exploring new model architectures that can better support these refined data, as well as investigating how intelligent agents can collaborate to achieve self-evolution with enhanced individual data. This is based on our belief that second-phase average models will continue to strengthen, providing a foundation for future-oriented designs.


\subsection{Principles to Achieve Model Self-Evolution}
The ability of a model to self-evolve is crucial for its long-term adaptability and personalization, and this depends heavily on an effective memory mechanism. In this context, we propose that long-term memory (LTM) provides the historical data accumulation and experiential learning capacity necessary for continuous model evolution. Just as humans refine their cognition and behavior through experience and memory, LTM enables a model to gradually optimize its reasoning and learning capabilities when dealing with long-term, dispersed, and personalized data.

\subsubsection{Empower Foundation Models with LTM Data for Self-Evolution}

In traditional LLMs, updating the model typically requires adjustments to all parameters, which is impractical for processing individual-specific data~\cite{roberts2020knowledgepackparameterslanguage}. A more optimal approach is to employ localized updates, allowing the model to adapt to sparse~\cite{NEURIPS2023_44956951}, personalized LTM data without compromising the stability of the global model. This method addresses the issue of individual data being ``averaged out'' in current models, enabling more comprehensive expression of personalized information.

Techniques such as Retrieval-Augmented Generation (RAG) with In-Context Learning (ICL) and Low-Rank Adaptation (LoRA) for fine-tuning (SFT) can be seen as ways to locally update individual data. We have developed a mixed strategy to integrate LTM data, yielding promising results in practical applications. However, this may not a perfect solution, and we are continuing to explore how to effectively integrate long-tail individual data into the model's memory mechanisms, hoping to attract more researchers to contribute to this field of exploration.

\subsubsection{Real-Time Weight Updates Combined With LTM Data for Self-Evolution}

Current LLMs typically separate the inference and training phases, where model weights are frozen during inference, preventing adjustments and learning based on new input~\cite{Lu2022}. This fixed inference process limits the model’s adaptability, particularly when handling personalized tasks and real-time learning. Inspired by the human brain’s updating mechanism, we believe that future LLMs should integrate inference and training with LTM, enabling the model to dynamically adjust weights upon receiving new information, akin to the continuous learning ability of humans. We also provided an overview of early work in the following sections, demonstrating how the integration of real-time training and inference allows LLMs to become more flexible and quickly adapt to new tasks or long-tail data. Additionally, this integration could help the model self-reflect and correct faulty reasoning paths when faced with complex inference tasks, improving both accuracy and efficiency. This dynamic self-adjustment capability would greatly enhance the model's personalization capacity and its potential for long-term evolution.

With LTM, a model can not only learn from short-term memory but also extract valuable insights from historical data, forming a deeper understanding of individual preferences and behavior patterns over time~\cite{Ballarini2009}. This understanding lays a solid foundation for personalized customization and dynamic adjustments, allowing the model to evolve more effectively. Especially when faced with new or extreme situations, LTM enables the model to reference past experiences, quickly make adjustments, and self-evolve, thereby achieving greater flexibility and adaptability.

\subsection{The Implementation Path of LTM in Model Self-Evolution}

Inspired by the importance of LTM for humans, we argue that research into long-term memory is essential for advancing model personalization. As AI models and intelligent agents continue to evolve, their foundational capabilities, akin to an increase in machine intelligence ``brain capacity'', provide greater support for the integration of long-term memory into personalized models. While efforts have been made to construct memories or experiences for evolving AI systems, significant gaps remain in defining, constructing, and evaluating long-term memory for AI, hindering the development of personalized LLMs. We begin by defining AI self-evolution and LTM, exploring the key role of LTM within it. We then focus on how LTM can be utilized to enable AI self-evolution. Our research focuses on three questions:
\begin{enumerate} 
    \item \textbf{What is AI self-evolution, and what constitutes long-term memory?} Why does AI self-evolution require models to have personalized capabilities? Why is long-term memory essential for achieving true personalization? What are the shortcomings of memory mechanisms in current language models, and how can these deficiencies help us refine the definition of LTM? 
    \item \textbf{How to construct LTM for self-evolution?} What types of data are most suitable for forming the foundation of a model’s personalized long-term memory, and how can we distill and structure this raw data into LTM? 
    \item \textbf{How to use LTM for AI self-evolution?} How can we efficiently process and utilize individual data to continuously update long-term memory, ensuring that it not only understands individual preferences but also enables self-evolution by adapting and coordinating with the environment as data grows and changes? \end{enumerate}

The main contributions of our study are summarized as follows:

\begin{itemize}
    \item \textbf{Definitions of AI Self-Evolution and LTM.} We provide an in-depth discussion of the relationship between AI self-evolution and LTM, proposing a systematic framework that highlights the core role of LTM in the process of AI evolution. Through LTM, AI not only addresses personalized needs but also continuously learns and optimizes, bridging the gap between general models and truly personalized intelligent systems. By effectively handling individual long-tail data, the long-term memory mechanism significantly enhances the individual capabilities and diversity of agents, laying the foundation for AI self-evolution. 
    \item \textbf{Data Framework for LTM.} To implement long-term memory, we developed a data collection, analysis, and synthesis framework that allows for differentiated system deployment based on various business scenarios. To verify the generalization ability of this framework, we deployed independent data systems in two distinct business contexts—office collaboration and health management. Specialized intelligent agents collaborate by integrating data from individual sub-models into a unified long-term memory.Each agent focuses on specific aspects of the data, ensuring seamless and accurate personalization even when data is sparse or inconsistent. Based on this data framework, we successfully established the world’s largest real-user voice dataset for mental health (see Section~\ref{chap7: data collection smhc1}), and augmented it through data synthesis (see Section~\ref{chap7:data_synthetic_memtal}). We are planning to open this dataset on a data platform to further support scientific research.
    \item  \textbf{Development Framework for LTM.} We propose a multi-agent collaborative development framework (OMNE) based on LTM. In this framework, each agent has an independent system structure that allows for autonomous learning and storing of a complete world model, thereby constructing an independent understanding of the environment. Through this LTM-based collaborative development, AI systems can adapt in real time to changes in individual behavior, optimizing task planning and execution, further promoting personalized and efficient AI self-evolution. We detail this framework in Section~\ref{subsec:omne}, demonstrating its potential in leveraging individual data for decision-making.
\end{itemize}

Our research contributes both theoretically and practically by integrating LTM into model personalization to promote AI self-evolution, with progress already made in practical applications. First, we discuss the importance of AI self-evolution and the critical role of model personalization in Section ~\ref{subsec:main_why_mp}. Next, we examine memory mechanisms in current LLMs and humans, exploring how human memory systems can inspire the design of LTM for model personalization, followed by a definition of LTM in Section \ref{subsec:main_ltm_mp}. The questions of how to construct LTM (Question 2) and How can LTM be used to achieve model personalization in AI Self-Evolution(Question 3) are addressed in detail in Sections \ref{subsec:how_to_LTM} and \ref{subsec:how_to_integrate} , respectively. Our efforts and results are presented in Section \ref{subsec:our_practice}, with further discussions and conclusions summarized in Sections \ref{subsec:main_feature_plan} and \ref{subsec:main_conclusion}.

\section{AI Self-Evolution}
\label{subsec:main_why_mp} 
The process of AI self-evolution can be compared to the Thousand Brains Theory or biological individual evolution. In the Thousand Brains Theory proposed by Jeff Hawkins \cite{hawkins2021thousand}, the brain does not operate through a single centralized system; rather, it constructs an understanding of the world through thousands of mini-models in the neocortex. These mini-models function independently while working together to form a diverse, distributed intelligence system. This theory challenges the traditional linear understanding of the brain by emphasizing that intelligence arises from the parallel processing and collaboration of multiple models. The Thousand Brains Theory posits that each region of the brain can independently create maps of the world and interact with maps from other regions, resulting in a more accurate and comprehensive cognition. Therefore, intelligence evolves progressively through the collaboration of multiple independent models.

On the other hand, the history of biological evolution demonstrates that there is no single “superorganism” dominating ecosystems \cite{johnson2010deconstructing, HAGENS2020106520}. Instead, the diversity driven by individual adaptations and mutations has enabled the formation and flourishing of the complex network of life we observe today.

Similarly, AI self-evolution can follow a path of multi-agent collaboration. In a multi-agent system, different agents interact, learn, and collaborate with one another to optimize their capabilities, generating personalized data. This personalized data serves as the foundation for continuous AI evolution, driving it from an initial general-purpose model to a system that increasingly adapts to individual needs. Just as biological evolution forms complex ecosystems through mutation and adaptation, AI self-evolution relies on diversity and co-evolution. This evolutionary path enables AI to continuously adapt to different environments and requirements, forming a more diverse and flexible intelligence system.

In the following sections, we will define AI self-evolution, break down the key system dependencies within AI self-evolution, and discuss how these capabilities contribute to more effective and adaptive self-evolution.

\subsection{Definition of AI Self-Evolution}
\textbf{Definition: AI self-evolution refers to the process by which AI models achieve breakthroughs in multi-agent collaboration and cognition through continuous learning and optimization with personalized data. This process is based on a shared core architecture, where each model evolves by processing personalized experiences and data, thereby enhancing its reasoning capabilities and adaptability, ultimately achieving autonomous learning and continuous evolution in dynamic environments.}

Model self-evolution enables AI models to continuously learn from personalized data, adapting to ever-changing environments and meeting diverse needs without relying heavily on human intervention. Throughout this process, models process and absorb new experiences, optimizing their architecture and outputs, evolving from generalized knowledge to more contextually adaptive personalized knowledge. Through dynamic learning mechanisms, models can retain and utilize key information from past interactions, supporting future decision-making and effectively mitigating issues like overfitting and data drift.

A key feature of model self-evolution is that it is based on a unified foundational architecture, ensuring that all model instances share a consistent core structure. However, the evolution of each model is driven by the unique experiences and data it processes, with differences between models arising from their individualized handling of personalized data. This approach ensures that, while models adhere to consistent internal rules and mechanisms, they can develop in differentiated ways according to personalized needs and environments. As evolution progresses, models become better at simulating individual behaviors, providing personalized and precise context-aware outputs, ultimately laying a solid foundation for multi-agent collaboration and cognitive breakthroughs.

\subsection{Key System Dependencies for AI Self-Evolution}
The realization of AI self-evolution does not happen spontaneously; it relies on a series of key system dependencies that provide the necessary foundation and framework for AI models to learn, optimize, and evolve into more personalized agents in ever-changing environments. These dependencies include not only the mechanisms of multi-agent collaboration but also the generation of personalized data, the construction of long-term memory, distributed model updating mechanisms, and self-correction mechanisms. These interdependent factors collectively drive the transition of AI from static models to self-evolving systems, helping AI transcend the limitations of traditional intelligence and gradually move toward the future of AI self-evolution.

\subsubsection{Multi-Agent Collaboration Mechanism}
The multi-agent collaboration mechanism is a key element of AI self-evolution, especially in handling complex tasks where efficient cooperation among multiple agents can significantly enhance the overall system performance\cite{Li_2023}. With the rise of large language models (LLMs) demonstrating the phenomenon of "emergent intelligence," the capabilities of AI systems have been greatly enhanced, accelerating the development of AI models\cite{NEURIPS2023_adc98a26}. These LLMs, with their vast number of parameters, endow agents with stronger memory, reasoning, and adaptability, allowing AI to perform remarkably well in more complex tasks. However, from the perspective of multi-agent collaboration, model personalization becomes the core factor promoting agent collaboration and evolution, particularly in solving complex problems and tasks.

Just as in human intelligence evolution, where increased brain capacity enhanced memory, thinking, and reasoning abilities, propelling the development of civilization\cite{hewer2012history}, LLMs similarly drive leaps in AI capabilities. However, current research suggests that in more complex scientific and engineering problems, human collaboration is essential. Similarly, whether AI systems will experience a second wave of emergent intelligence in the future depends on whether collaboration among multiple agents can be elevated to a new level.

Some explorations suggest that in small-scale multi-agent collaborations, increasing the number of agents does bring some performance improvements, but these gains are not consistent or stable\cite{oroojlooy2023review}. We believe the key issue behind this lies in the fact that most current multi-agent collaborations are still limited to role-playing interactions, where the agents' capabilities and knowledge are often homogeneous, lacking the differentiated skills required for deep collaboration. To achieve significant breakthroughs in large-scale multi-agent collaboration, it is crucial to rely on agent personalization, which can provide each agent with unique expertise and abilities, thus promoting efficient collaboration and evolution of the system.

Therefore, the true potential of multi-agent collaboration depends on the creation of a group of differentiated, highly personalized agents. This not only supports the resolution of complex tasks but also provides a new path for the self-evolution of AI systems. Model personalization is an indispensable part of this evolutionary process, enabling different agents to contribute uniquely in collaborations, pushing AI systems toward higher levels of intelligence.

\subsubsection{Differentiated Personalized Models}
Personalized data generation is one of the core driving forces behind AI self-evolution, especially in multi-agent systems where having personalized models for each agent is crucial for ensuring system diversity and efficient collaboration. As AI application scenarios become increasingly complex, models must continuously acquire, process, and respond to personalized data to dynamically adjust to the needs and preferences of different individuals, truly meeting diverse task requirements\cite{zha2023data}. Each agent, through its personalized model, can not only handle specific tasks independently but also contribute unique insights and experiences during collaboration with other agents, thereby generating more diverse and personalized data within the overall system.

This diversity brings stronger collaborative capabilities to AI systems. Through effective interactions among agents, personalized models can better respond to the needs of different individuals and support continuous model evolution through the accumulation of long-term data and continuous learning. In complex fields like healthcare\cite{delpierre2023precision}, the requirement for model personalization becomes especially prominent when handling multimodal and heterogeneous data. A one-size-fits-all strategy shows limitations in addressing these complex tasks, while personalized models with differentiated processing capabilities can dynamically adapt to individualized scenarios, delivering precise and efficient performance.

Moreover, even within the same task scenario, different individuals will have significantly different expectations of the model\cite{Bandi2023ThePO}. Through personalized models, AI systems can dynamically adjust according to these varied needs, providing highly tailored services to each individual. For example, in dialogue systems, individuals have unique preferences for the style, tone, and even response format of the model’s output. To achieve precise personalized services, models must have dynamic learning capabilities, deepening their understanding of individual needs through continuous interaction.

Therefore, personalized models not only promote the diversity and collaboration of AI systems but also support the self-optimization and evolution of the entire system by generating more personalized data. This process forms a positive feedback loop: through the collaborative work of multiple agents, AI systems generate increasingly personalized feedback, continuously meeting individual needs and further enhancing model capabilities, ultimately achieving true self-evolution.

\begin{figure}[ht]
    \centering
    \includegraphics[width=0.8\textwidth]{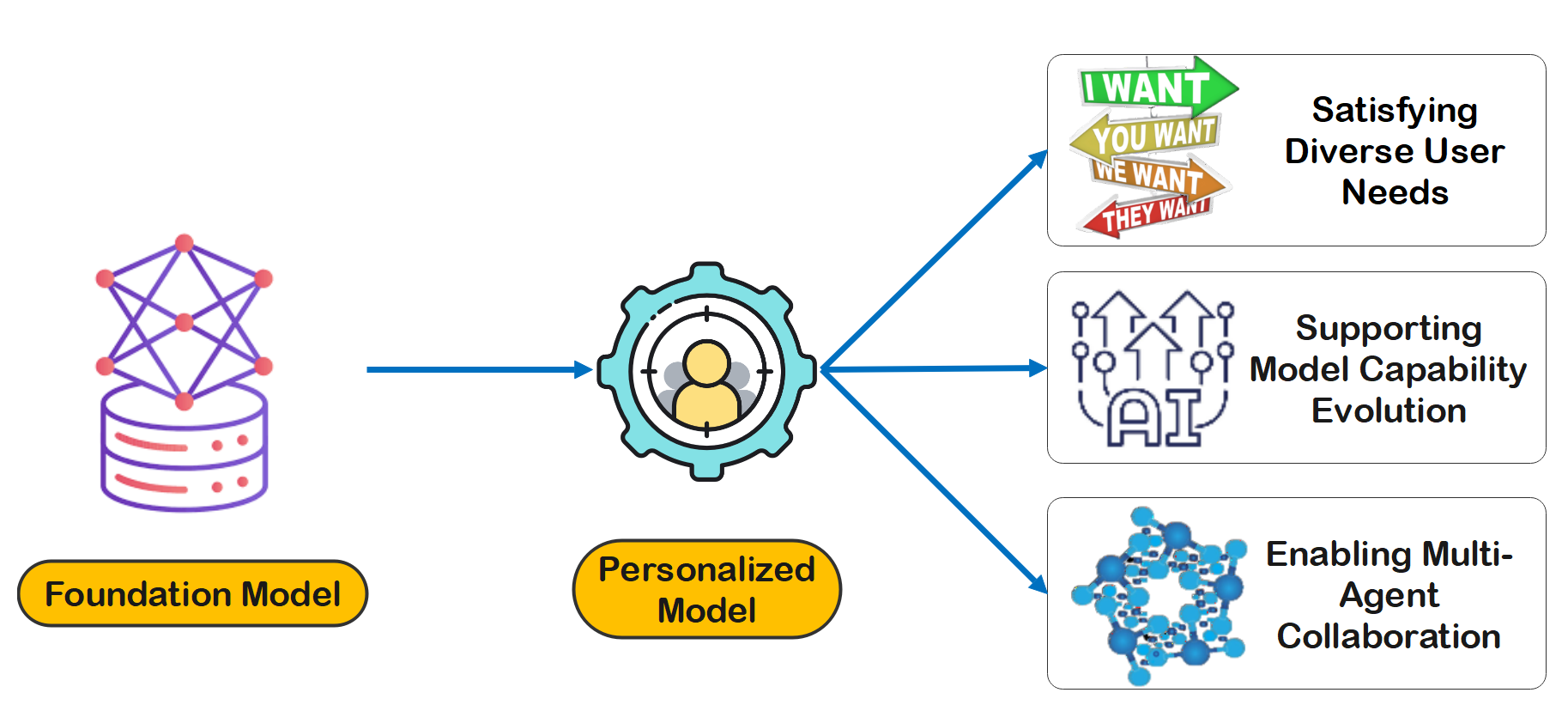}
    \caption{Model personlization in AI self-evolution}
    \label{fig:sec2_model_personalization}
\end{figure}

\subsubsection{Self-Correction and Evolution Mechanism}
To achieve AI self-evolution, models must possess a self-correction mechanism \cite{talvitie2017self}. This mechanism not only ensures the internal consistency of the model but also enables it to adapt to changes in the external environment. Self-correction is at the core of AI self-evolution, allowing models to update their cognition and behavioral strategies through continuous feedback loops. This is similar to the process of biological evolution, where selection and adaptation drive continuous optimization to fit the environment.

Recent mathematical studies suggest that predictable information often exists in the low-dimensional structures of high-dimensional data \cite{schon2023} \cite{NIPS2017_c7558e9d} \cite{SMITH199250}. Understanding and leveraging this concept is crucial for AI self-evolution. By identifying the low-dimensional structures within data, AI can learn and generalize more effectively, thus enhancing its autonomous learning and evolutionary capabilities.

\subsubsection{Long-Term Memory and Learning Ability}
Long-Term Memory (LTM) is a fundamental cornerstone in the process of AI self-evolution, providing models with the ability to accumulate historical experiences and knowledge, enabling continuous optimization through long-term interaction and learning. LTM can store data not only at the individual level but also accumulate data over time, helping models adjust their responses and behaviors based on this information, thus facilitating self-evolution. Compared to most personalized approaches that rely on a context window, long-term memory overcomes the limitations of short-term approaches by endowing models with the ability for continuous learning and self-improvement, enabling them to exhibit stronger adaptability when faced with complex environments and multi-agent collaboration.

Just as humans use long-term memory to shape their behavior and identity, AI systems can also utilize LTM to provide customized responses based on individual data. This “individual data” is not limited to user interaction data but can also include the specific needs of an organization or domain, allowing the model, through long-term accumulation, to surpass the framework of general knowledge and support more precise decision-making and behavioral adjustments. Therefore, LTM is crucial for enabling AI to achieve continuous self-evolution.

This paper emphasizes the role of long-term memory as a core driver of AI self-evolution. To truly realize AI self-evolution, we need models equipped with LTM, which can store and manage long-term interaction data in real-world environments and effectively represent long-tail data from individuals. Through this expression of diversity, models can continuously adjust themselves in collaboration with the environment and multiple agents, thus promoting the process of self-evolution. As such, long-term memory is not only the foundation for personalized data generation but also a key mechanism supporting the long-term adaptation and optimization of AI systems. In subsequent chapters, we will discuss the critical role of long-term memory in AI self-evolution in greater detail.

\subsection{Thought Experiment: From Euclidean Geometry to Riemannian Geometry}
An interesting question is: \textit{Can a large language model deduce Riemannian geometry from the five axioms of Euclidean geometry?} This would not only require the model to perform logical deductions based on existing axioms but also to examine and challenge fundamental assumptions—such as the parallel postulate—thereby stepping into the entirely new domain of non-Euclidean geometry. Current LLMs are not yet capable of such creative leaps, as they rely on existing data for reasoning and lack the ability to propose new hypotheses and extend the boundaries of knowledge.

Suppose LLMs could continuously adapt and update with personalized data, gradually developing sensitivity to atypical patterns—could they eventually overcome this limitation? In this thought experiment, the volume of data is not a constraint, as synthetic data could be continuously generated. If the model could not only process human-provided data efficiently but also reflect on its own knowledge, dynamically adjust, and propose new hypotheses, using experimental deduction to validate them, then such a system would no longer be just a passive tool for knowledge storage and reasoning. Instead, it would become a cognitive entity capable of self-evolution, breaking through established mathematical frameworks and pioneering new theoretical realms.

From this perspective, the potential of real-time updates and personalized learning lies not only in improving the accuracy of existing knowledge processing but also in granting the model higher levels of cognitive flexibility. This would enable it to genuinely participate in the discovery of new knowledge. Realizing this potential may be the key to overcoming the current limitations of LLMs, evolving them from data-driven intelligence into explorers capable of expanding the boundaries of knowledge.

\subsection{Future Directions and Challenges}
As explored in this chapter, personalized models lay the foundation for multi-agent collaboration and cognitive breakthroughs in increasingly complex task environments. By integrating reasoning with training, allowing local updates, and adopting distributed architectures, AI will not only simulate human language outputs but also develop human-like reasoning and innovation abilities. Ultimately, AI will break through current technological limitations by interacting with the physical world, autonomously learning, and continuously evolving, driving the expansion of cognitive boundaries.

While the prospect of AI self-evolution is promising, there remain significant technical and theoretical challenges in achieving this goal. After establishing the foundation for self-evolution through model personalization, key challenges for AI self-evolution include: \begin{itemize} \item How can models be effectively evaluated and achieve self-evolution? \item How can collaboration mechanisms among agents be designed? \item How can we break through the current Scaling Law of models and continuously improve performance? \end{itemize}

To address these questions, further discussions are presented in section \ref{subsec:main_feature_plan}. Our future research will focus on developing AI systems with autonomous learning, exploration, and evolutionary capabilities. In the next chapter, we will focus on the importance of long-term memory in AI self-evolution.

\section{LTM for AI Self-Evolution}
\label{subsec:main_ltm_mp}

AI self-evolution emphasizes the ability of AI systems to improve their capabilities through continuous learning and adaptation. In this process, the retention and updating of individual model information by standalone AI models is a crucial feature of the entire AI system's self-evolution. However, most current methods for implementing LTM in models primarily rely on context windows. These models typically often utilize immediate context or recent individual interactions to generate responses \cite{taoSurveySelfEvolutionLarge2024}. While this approach can achieve a certain degree of diversity, it has significant limitations in supporting long-term learning and continuous adaptation, hindering models from achieving true self-evolution. 

Therefore, in this chapter, we will explore in depth the crucial role that LTM plays in AI self-evolution. We first define LTM in the context of AI self-evolution and analyze the shortcomings of current LLM memory mechanisms. We then discuss enhancing AI models' self-evolution capabilities by drawing inspiration from human LTM characteristics and addressing the challenges and potential solutions in achieving this goal. Through these discussions, we aim to provide new ideas and directions for building AI systems capable of continuous learning and self-improvement.

\subsection{Definition of LTM in AI Self-Evolution}

\textbf{Definition: LTM is the information that can be retained and utilized by AI systems over extended periods, enabling models to adjust their responses and behaviors based on a broader context.}

Just as humans use LTM to shape their behavior and identity, AI systems can employ similar mechanisms to customize their responses and behaviors based on individual data. Here, ``individual data" is not limited to individual users but also includes specific organizations and domains, allowing models to adjust their responses and behaviors according to broader individual contexts and needs. This forms the basis for creating personalized models that go beyond mere general knowledge.

LTM can be understood as a vast and complex repository of refined knowledge, shaped over time by a group of independent yet harmoniously interacting agents—similar to cortical columns in the brain. Each agent functions as an autonomous unit capable of learning, refining, and storing a comprehensive model of its own corner of the world. However, these agents do not operate in isolation; they contribute their insights to the broader LTM collective, creating a shared knowledge base that supports deeply personalized interactions.

Unlike traditional static data storage systems, the LTM framework is a dynamic and distributed memory framework, akin to a network of independently operating thoughts in the human brain, where insights from various independent learning processes can merge. Just as a society comes together to form a more coherent understanding, this collective intelligence enables the system not only to accumulate knowledge but also to synthesize it in ways that better reflect the complexity and nuances of user needs, ultimately achieving AI system self-evolution. LTM demonstrates a more nuanced and comprehensive understanding of both individuals and collectives, enabling the system to respond to personal needs with a level of granularity that reflects this complexity. In this sense, LTM transcends mere data storage—it becomes an adaptive, continuously evolving cognitive organism, constantly refining itself in response to its environment, much like the human cognition it seeks to emulate.

\subsection{Limitations of Current LLM Memory Mechanisms}

LLMs like GPT-4 and Gemini demonstrate advanced intelligence and a comprehensive understanding of the world. However, to achieve true self-evolution, these models must be able to effectively process, store, and integrate information acquired through continuous interaction with various environments. Currently, LLMs primarily manage information through two memory mechanisms: contextual memory and compression-based parametric memory. While these mechanisms perform excellently in short-term tasks, they still fall short in supporting long-term autonomous learning and evolution.

\subsubsection{Memory through Prompting}

Current LLMs utilize prompts as a form of contextual memory to retain and leverage information during inference. The prompt, which includes both the instruction and relevant context, serves as a temporary memory buffer, allowing the model to process and generate content based on the given context. This mechanism enables LLMs to perform a wide range of tasks without task-specific fine-tuning. However, from a self-evolution perspective, this prompt-based memory mechanism has several key limitations:

\begin{itemize}
     \item \textbf{Temporality:} Stored context information is limited to the scope of the current task. Once a single inference call is completed, the model discards its previous state and sequence information, meaning it cannot utilize previously acquired knowledge in subsequent tasks.
    
    \item \textbf{Absence of Continuous Learning:} Unlike systems with explicit memory mechanisms (e.g., Long Short-Term Memory (LSTM) networks \cite{hochreiter1997long}), LLMs do not have an intrinsic ability to accumulate and refine knowledge across multiple interactions or tasks based solely on prompts.
    
    \item \textbf{Limited Cross-Task Knowledge Integration:} While prompts can provide task-specific context, they do not facilitate the automatic integration of knowledge across diverse tasks or domains. This hinders the model's ability to develop a cohesive understanding that evolves over time.
    
    \item \textbf{Dependency on External Curation:} The quality and relevance of contextual information heavily rely on how prompts are crafted by users or systems. The model itself cannot autonomously curate or optimize its contextual knowledge base.
\end{itemize}

\subsubsection{Parametric Compressed Memory}

Another memory mechanism is compression-based parametric memory, which forms a type of LTM by compressing world knowledge into the model's parameters~\cite{brown2020fslearner}. This mechanism allows models to retain key information over longer periods, but it has two significant drawbacks:

\begin{itemize}
    \item \textbf{Unable to Update in Real-Time:} Since the model's parameters are fixed during the training process, LTM cannot be easily updated once training is complete. This limits the model's diversification and adaptability to the environment, as it cannot quickly learn to generate new memories to adjust its behavior.

    \item \textbf{Difficulty in Expressing Individual Data:} As statistical models, Transformers struggle to adequately represent individual data in their LTM. This typically leads to one of two outcomes: either the model fails to retain individual data, or it overfits and forgets previously-stored world knowledge, resulting in memory bias. The current approach to mitigate this issue is through incremental training with carefully balanced data proportions to incorporate individual data into the model. However, this process is inefficient and difficult to scale in large-scale or dynamic data environments.
\end{itemize}

These limitations indicate that AI models need a more flexible and adaptive memory system that can retain individual data and achieve real-time adjustments, similar to human LTM.

\subsection{Inspiration from Human LTM}
\label{subsec:def_HLTM}

To better understand LTM in AI systems and achieve AI system self-evolution, we can draw inspiration from the concept of human LTM. Existing neuroscience research suggests that personal memories generated through human interaction with the world are key factors in forming diverse and personalized behaviors. Current research typically divides human memory into three main types: working memory, short-term memory, and LTM. Working memory and short-term memory mainly contain temporary information related to current tasks or situations, which is quickly forgotten if not processed and transformed into LTM. Therefore, LTM can be considered the key data foundation for the formation of human personality.

Specifically, human LTM refers to the brain's ability to store and retrieve information over extended periods, ranging from hours to decades\cite{McGaugh2000}. Unlike short-term memory, which is temporarily used for immediate use, LTM is responsible for preserving knowledge and experiences that can influence our future behavior\cite{Squire2004,Sridhar2023}. This type of memory includes various subtypes, including episodic memory (personal experiences), semantic memory (general knowledge), and procedural memory (skills and habits)\cite{Merlo2024}.

The formation of LTM involves multiple processes, including encoding, consolidation, and retrieval\cite{Straube2012}. Encoding is the initial acquisition of information, while consolidation is the process of stabilizing new information and integrating it into existing memory networks\cite{Dudai2004}. Retrieval is the process of accessing and utilizing stored information when needed. These mechanisms are supported by neural processes in the brain, including the hippocampus and various cortical regions\cite{Squire2011}.

Moreover, LTM not only influences the formation of personal interests and habits but also plays a crucial role in the emergence of diverse needs\cite{sutin2023five}. Additionally, LTM significantly affects an individual's knowledge accumulation\cite{kenett2023thirst}, problem-solving abilities\cite{hambrick2003role}, social adaptability and self-regulation abilities\cite{zhao2011social}, leading to different expressions of intellectual development and evolution. In social life, social experiences and memories stored in LTM help individuals build trust, promote knowledge and resource sharing, and ultimately drive cooperation and collaboration\cite{weldon1997collective}.

\subsection{LTM in AI Models}

\begin{figure}[h]
    \centering
    \includegraphics[width=0.8\linewidth]{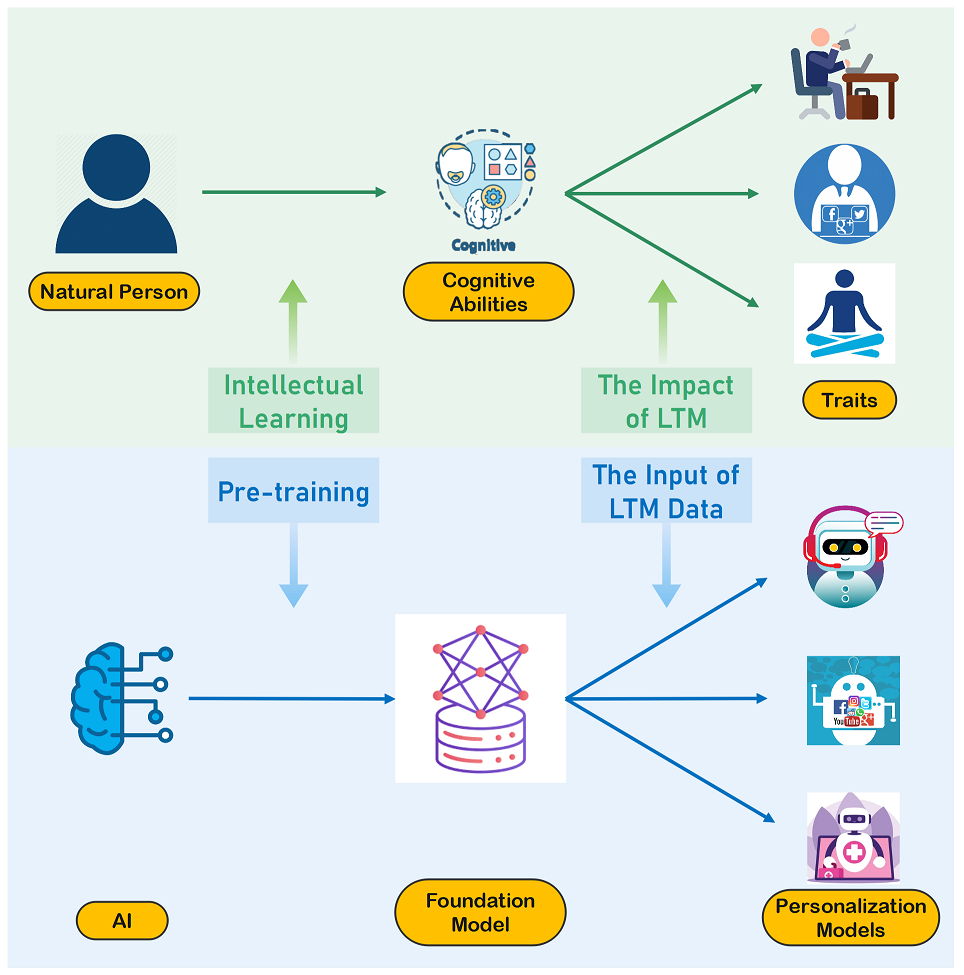}
    \caption{The difference between human learning and growth versus artificial intelligence training and evolution.}
\end{figure}

In the previous section, we emphasized the important role of LTM in human evolution, which relies on a comprehensive mechanism for memory formation, updating, and utilization to ensure LTM effectively serves humans. Many aspects of model design and development draw inspiration from human cognitive and reasoning structures, such as the propagation mechanism of neural networks. So if LTM can aid the progress of human society, can it also be used for AI self-evolution? Our answer is affirmative, and we discuss this from the following aspects:

\begin{itemize} 
    \item \textbf{From the perspective of data accumulation:} Both models and humans interact extensively with their environment, providing foundational data for personalization. Compared to humans, AI models can interact with their environment more efficiently and can perform these interactions and iterations in purely virtual, digital environments. Therefore, by designing appropriate memory refinement strategies, models should be able to accumulate long-term memories like humans, possibly even with higher efficiency and scale.
    
    \item \textbf{From the perspective of model updates:}
    Artificial intelligence excels in storing and calling upon vast amounts of data, far surpassing the scale of human memory. Neural networks manage this data through distributed parameters, processing inputs from different domains. However, this storage is relatively rigid, lacking the flexibility for real-time updates and typically requiring retraining to implement updates. In contrast, human memory is highly adaptive, quickly integrating new information and discarding outdated details through a process of "adaptive forgetting"\cite{Wimber2015}. This flexibility helps humans avoid cognitive overload and focus on the most relevant information at hand. To match this capability, AI systems need to develop dynamic update mechanisms that allow them to selectively update knowledge and discard outdated information without comprehensive retraining.

    \item \textbf{From the perspective of utilizing LTM:}
    Current advanced LLM memory mechanisms, such as contextual memory and parametric memory, can store and utilize large amounts of information with the advantage of non-forgetting. Moreover, AI can combine instant memory update mechanisms when encountering important new data, which is particularly useful in dynamic environments. This provides AI with certain advantages over human memory utilization, such as larger storage capacity and faster retrieval speed. However, these memory utilization methods still face significant challenges in managing dynamically accumulated long-term memories, especially in terms of flexibility and efficiency of memory updates, and still struggle to reach the level of human memory systems. Future research needs to focus on how to achieve efficient, flexible memory update and utilization mechanisms like humans.
\end{itemize}

In summary, the importance of human LTM to human society points the way for the development of AI systems. While human memory excels in context integration, flexibility, and real-time updates, AI performs better in managing large-scale datasets and identifying patterns. To fully leverage AI's advantages and address its shortcomings, we should combine AI's flexibility with its ability to process at scale, while introducing prioritization mechanisms and efficient information update methods similar to human memory. This hybrid approach can produce more personalized, responsive, and context-aware AI models, bringing them closer to the complexity and adaptability of human cognition.

\subsection{The Importance of LTM for AI Systems}

\subsubsection{LTM in Single Models} LTM is crucial for diversity and AI self-evolution because it allows models to generate new outputs based on a deep understanding of individual interactions. By recalling past interactions, identifying preferences, and adjusting context over time, LTM enables AI systems to infer new response patterns from historical data and provide adaptive answers in new environments. This capability is rooted in the information encoding within neural network weights and parameters, which is critical for supporting personalized interactions and continuous learning\cite{Zhong2024}.

The distributed and dynamic nature of LTM allows AI systems to optimize interactions based on historical data, individual preferences, and context\cite{Du2024}. Ultimately, integrating LTM into tens of thousands of AI models marks a significant step towards AI evolution. The data from LTM lays the foundation for personalized and context-aware interactions, evolving continuously over time. As AI technology continues to advance, the development of LTM will become key to creating systems that not only understand individual needs but can also anticipate and adapt to unique requirements, thus providing truly intelligent experiences.

\subsubsection{The Role of LTM in AI Self-Evolution} 

The ability for models to self-evolve is a highly promising direction in AI technology development and application. Just as human intelligence relies on personal interactions and feedback from the real environment, the LTM data accumulated by models during interactions provides the most critical data support for self-evolution. Of course, there are some differences:

Unlike human evolution, LTM-driven model evolution is not limited to real-world interactions. On one hand, models can interact with the physical environment and receive direct feedback like humans, which, after processing, will enhance their capabilities. This is also a key area of research in embodied AI. On the other hand, models can interact in virtual environments and accumulate LTM data, which has lower costs and higher efficiency compared to real-world interactions, thus achieving more effective capability enhancement.

Compared to existing methods of directly storing interaction history and reflection, refined LTM data can provide high-quality data support for model self-evolution, further accelerating the efficiency and effectiveness of model evolution. This provides more effective support for enhancing the capabilities and rapid application of large model (LLMs) driven agents.

Having clarified the importance of LTM for AI model self-evolution, we will focus on discussing the following two questions: 1) How to construct LTM (Section \ref{subsec:how_to_LTM}). 2) How to integrate LTM into models (Section \ref{subsec:how_to_integrate}).





\section{How to Construct LTM?}
\label{subsec:how_to_LTM}
The construction of LTM represents a critical aspect in model personalization. Raw data serves as the foundation and source of LTM. Therefore, the relationship between LTM and raw data will be explored in the following sections. Subsequently, the method of refining raw data into LTM will be introduced, culminating in a review of existing methods to get high quality data for LTM, and he discussion consists of two main parts: data collection and data synthesis.

\begin{figure}[ht]
    \centering
    \includegraphics[width=0.9\linewidth]{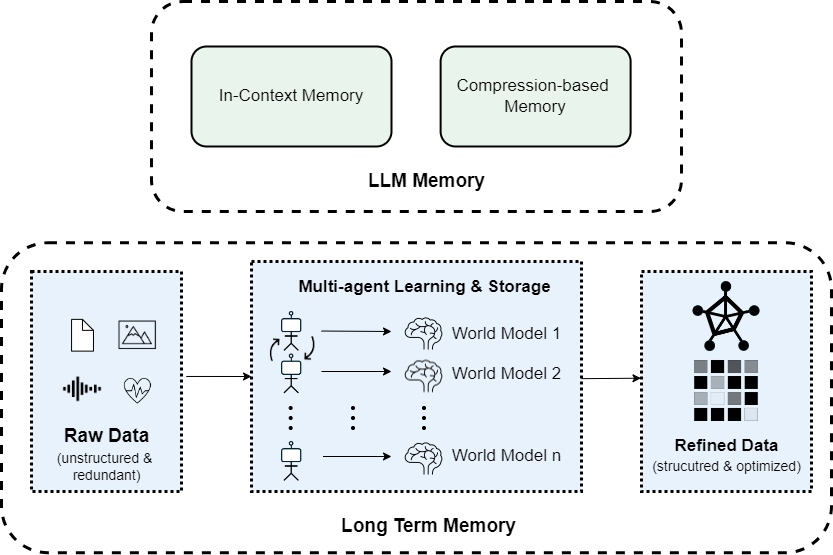}
    \caption{Comparison of LLM Memory and Long-Term Memory}
    \label{fig:ltm_vs_llm_memory}
\end{figure}

\subsection{Raw Data to LTM}

Raw Data represents the comprehensive collection of all unprocessed data that a model receives through interactions with the external environment or during the training process. This data includes a wide range of observations and records, which may contain both valuable patterns and large amounts of redundant or irrelevant information. While Raw Data forms the foundational layer for the model's memory and cognition, it requires further processing to be effectively used for personalization or efficient task execution.

LTM refines and structures this raw data, making it usable by the model. This process enhances the model's ability to deliver personalized responses and recommendations. While Raw Data captures direct observations that may result in redundancy and clutter, LTM organizes this data into a structured memory, enabling the model to recognize patterns, remember preferences, and provide adaptive responses.

For instance, in an AI medical scenario, Raw Data might capture basic patient information, such as demographic details, current symptoms, and immediate diagnoses. However, Raw Data alone would struggle to effectively manage and integrate a patient’s comprehensive health history into a cohesive understanding of their condition.

In contrast, LTM refines and organizes this raw patient data into an interconnected structure, allowing the AI model to draw meaningful inferences from the patient’s entire medical history. This supports advanced reasoning and personalized healthcare solutions. LTM would not only retain a patient’s ongoing medication regimen but also relate it to their historical response to similar treatments, recognize patterns in evolving symptoms, and adapt recommendations accordingly. This refined memory enables the AI model to facilitate advanced reasoning and personalized healthcare that evolves over time, delivering deeply individualized care.

In conclusion, while Raw Data provides a foundation for immediate context management, LTM is indispensable for achieving the deep, sustained personalization necessary for AI systems to remain relevant and effective. This allows AI to deliver richer, more personalized experiences that continuously improve and adapt to real-world complexities.

\subsection{Construction Strategies of Raw Data}
\label{sec:issue_data}
Unlike the human brain that seamlessly receives and processes vast arrays of sensory inputs intoa memory,AI models face significant challenges in utilizing raw data to build LTM effectively. Critical issues that need to be addressed include:

\begin{itemize} 
    \item \textbf{Data diversity and representativeness: } 
   LTM must be diverse and representative to ensure model robustness across various scenarios and user groups. Imbalances and biases in collected data can lead to suboptimal performance. Additionally, the quality and consistency of data labeling are critical; inaccuracies can diminish model performance and personalization outcomes \cite{wang2014latent}.

    \item \textbf{User behavior capture and reasoning: } To achieve true personalization, the model needs to deeply understand user behavior patterns and reasoning processes. However, current LTM models struggle to capture intermediate reasoning steps and time-series information. This deficiency makes the model fully simulate human cognitive processes, resulting in outputs that may lack coherence and contextual relevance.
    
    \item \textbf{Data privacy and security: }The development of personalized LTM models has brought data privacy and security issues to the forefront, especially in on-device fine-tuning. Ensuring the security of data handling and processing is essential \cite{wei2024trustworthy}.

\end{itemize}

\subsubsection{Data collection framework}

Establishing a robust data collection and labeling process is essential for ensuring diversity and high quality LTM. The collection of data for personalized models begins with understanding the personal experiences, encompassing both digital footprints and physical interactions. Given the fragmented nature of data collected compared to the continuous, rich data processed by the human brain, it is crucial to employ advanced methods for both realms. We have developed a comprehensive framework for data collection, analysis, and synthesis, which allows for differentiated system deployment tailored to various business scenarios. To validate the generalizability of this framework, we deployed independent data systems in two distinct business contexts: office collaboration and health management. A detailed description of these data systems is provided in Sec\ref{chap7: data collection smhc}..

In the digital realm, human activities and behavior on various platforms form significant digital footprints, providing crucial insights into preferences, interests, and interaction patterns. Digital communication data, including messages, emails, and voice recordings, reveal language preferences, communication styles, and social connections \cite{huang2024levels}. Analysis of this data is essential for developing personalized AI systems capable of tailoring responses and enhancing human interaction experiences.

Behavioral data, such as web browsing histories, app usage patterns, and social media activities, offer a detailed perspective on human interests and behavior in digital spaces \cite{Cao2020Behavior}. This information is crucial for understanding how individuals interact with various digital environments, facilitating the development of more accurate personalized models. Content consumption data, including the types and genres of articles, videos, and music consumed, contribute to building comprehensive profiles of user preferences and trends over time. Research by Jeung and Huang emphasizes the significance of this data in facilitating tailored interactions and fostering personalized experiences \cite{jeung_huang2024unlocking}.

In the physical realm, wearable technologies and ambient sensors provide extensive data about human interactions with the environment. Devices such as AR glasses (e.g., Apple Vision Pro) capture continuous streams of visual and auditory data, reflecting environmental interactions. These devices are crucial for gathering comprehensive data inputs that are otherwise difficult to obtain through traditional methods. Recording personal interaction logs across various devices and contexts, such as smartphone usage and smart speaker commands, provides valuable insights into behavior. This practicefacilitates the integration and analysis of diverse data sources, improving the reliability and accuracy of personalized models \cite{Abb2022Reference}. Biometric signals, including heart rate, galvanic skin response, and brain wave activity, offer insights into human states and preferences. Biometric authentication systems based on signals such as ECG and EMG provide robust datasets for personalized modeling applications, enhancing the accuracy and reliability of AI systems \cite{Pradhan2022Open, Samarin2019Key}.

Contextual and behavioral data collection is crucial for a comprehensive understanding of human experiences. Such data includes location-based information and interactions within specific contexts, which provides a richer and more nuanced picture of human behavior. Tracking location data and correlating it with behavior patterns introduces essential spatial and temporal dimensions to the dataset. Hybrid location privacy schemes protect data privacy while maintaining the accuracy of services \cite{Sabir2022Hybrid}. 

To enhance the fidelity and richness of data collected for personalized models, future research should focus on Enhanced Sensory Integration and Context-Aware Data Collection. Leveraging advanced sensor technologies to capture richer, multi-modal data streams that closely mimic human sensory experiences is vital for improving data comprehensiveness and contextual richness. Developing advanced algorithms that can infer and fill contextual gaps in recorded data provides a more comprehensive understanding of personal experiences, thereby  leading to better predictions and personalization.

However, the diversity, representativeness, and quality control of the data are critical issues that must be emphasized during the collection process. High-quality personalized models require diverse and high-quality data to ensure robustness across different scenarios and user groups. Simultaneously, data privacy and security issues are paramount, and it is essential to protect user data during collection, storage, and processing.





\subsubsection{Data synthesis techniques}
Given the privacy concerns, limitations in individual data collection, and the fragmented nature of such data, synthetic data generation becomes increasingly critical. By producing synthetic data that closely mirrors real-world scenarios, we can compensate for the shortcomings of real data, providing rich context and diverse experiences for personalized modeling, thereby enhancing the relevance and practicality of the model.

Early approaches, such as utilizing human interactions and role-playing, were effective in producing high-quality datasets. For instance, one study \cite{Yao2022D4} introduced a three-phase human role-playing paradigm to synthesize a depression diagnosis dialogue dataset. However, human-involved methods still face significant challenges, including high costs and privacy concerns \cite{Kurakin2023Harnessing}. Additionally, other studies \cite{Singh2023Beyond, Gilardi2023ChatGPT} have emphasized that human-generated data may not be optimal for LTM construction due to inherent biases. 

Recent advancements in large language models (LLMs) offer a scalable and efficient alternative. Synthesizing data that mirrors real-world scenarios using LLMs can significantly enhance the pertinence and utility of LTM in personalized models. An iterative dataset synthesis method was proposed, in which large language models (LLMs) generate datasets for smaller models by progressively refining inaccuracies identified during the evaluation of the smaller models against real-world data. This approach is called Synthesis Step-by-Step (S3)\cite{wang2023synthesize}. It iteratively prompts the LLM to extrapolate errors encountered by smaller models in validation datasets, gradually narrowing the gap between synthetic and real data. Experimental results indicate that the S3 improves model performance across multiple natural language processing (NLP) tasks, with gains of up to 15.17\%, while reducing reliance on human-annotated data. This highlights its potential in enhancing model accuracy and reducing data annotation costs.

Simulating context-rich interactions is crucial for capturing the nuances of individual experiences. The construction of multi-agent systems, where agents assume various roles within text-based role-playing environments, offers an effective approach for generating diverse interaction data. For instance, the thespian agent framework offers a novel approach for emulating multiple characters in text role-playing games, facilitating the capture of diverse experiential nuances \cite{cui2023thespian}. This framework enables agents to play multiple characters using a soft-prompt mechanism, which guides the agent on which character to simulate. Through an integrated attention mechanism, agents learn new characters based on previously actions and feedbacks, enabling few-shot learning and enhancing adaptability in simulated environments. Experiments have demonstrated that this approach outperforms existing multi-character learning frameworks, making it highly effective for generating context-rich interactions that support LTM in personalized models.

Interactive simulations in dynamic environments is also essential for emulating the complexity of human experiences. The EnvGen framework employs LLMs to generate and adapt training environments tailored to the specific weaknesses of reinforcement learning (RL) agents, continuously adjusting configurations based on agent feedback \cite{zala2024envgen}. This dynamic adjustment fosters the development of robust and adaptable skills, outperforming static curriculum learning and generating synthetic data that closely mimics real-world interactions.

Enhancing agent-environment interactions is another promising approach for synthetic data generation. The Affordable Generative Agents (AGA) framework proposes cost-effective interaction paradigms to optimize agent-environment and inter-agent interactions\cite{yu2024affordable}. To achieve this, AGA substitutes repeated LLM inferences with learned policies for agent-environment interactions and compresses auxiliary dialogue information for inter-agent interactions. This reduces the computational costs while maintaining the quality of interactions. Extensive experiments have demonstrated the efficiency and effectiveness of AGA in producing believable low-cost interactions, making it an essential approach for scaling LTM synthesis in personalized models.

Given the inherent limitations of any single approach, a mixed data synthesis strategy offers a more robust and comprehensive solution. A holistic and flexible combined approach that incorporates multiple data synthesis techniques can yield superior results. The integration of real data synthesis with iterative error correction, role-playing adaptive dialogues, dynamic simulation environments, and agent-based interaction frameworks collectively forms a robust methodology for LTM in personalized models. This mixed-method framework leverages the strengths of each technique while mitigating their individual limitations, thus rendering a comprehensive and rich personalized model that can process a wide array of personal experiences effectively. In \ref{chap7.1.3: data generation RTG}, a mix-method data generation algroithm called: RTG synthesis method will be introduced, demonstrating its effectiveness in enhancing the model's performance in memory recall tasks without negatively impacting its overall capabilities.

\subsection{Construction Strategies of LTM}
\label{sec:coll_of_LTM}
LTM is the effective organization and structuring of raw data, not merely classifying and sorting raw data on the surface to make the information more organized. Instead, it is about designing and optimizing the forms of LTM storage from the perspective of rapid storage and retrieval of memory and efficient utilization of information. By establishing links between related information, effectively processing the data, and reorganizing the information, an intelligent agent can quickly locate the needed memory fragments, thereby improving response speed and accuracy. The following are several major operational modalities and their corresponding transformation methods.

\begin{itemize}
    \item \textbf{Text summarization:} Summary storage compresses long contexts into short, concise data forms. It involves inductively summarizing continuous interaction data, allowing LTM to exist in continuous and abstract forms. This method significantly reduces storage space through high abstraction and data compression, thereby enhancing retrieval efficiency. Summary storage is particularly suitable for scenarios requiring frequent retrieval and rapid understanding. For instance, OpenAI's ChatGPT employs this method by recognizing and summarizing interaction records in real-time, thereby generating short contextual memories aligned with the user's personal habits, which enhances its ability to meet personalized user needs during subsequent use.

    \item \textbf{Data structuring:} Structured storage predefines specific data structures, such as hierarchical structures, tree structures, key-value pairs, etc., to organize and store processed raw data. This approach makes information highly systematic, facilitating data retrieval and management. Moreover, it allows interaction with the stored data via database query languages like SQL, achieving more precise memory retrieval while also providing a bridge for multi-turn interactions between memory and intelligence.

    \item \textbf{Graph representation:} Graph-based storage uses nodes and edges to represent and manage information, boasting flexibility and scalability, which is suitable for representing the associativity of long-term memory content. Transforming raw data into a graph-based long-term memory requires data preprocessing, including data cleaning and format conversion, followed by identifying key entities and their relationships within the data to define the graph's nodes and edges. During the graph construction stage, these nodes and edges are added to the graph structure and optimized techniques such as graph sparsification and edge weighting are applied to enhance the storage and retrieval efficiency of the graph. These optimized graphs are stored in specialized graph databases like Neo4j to achieve efficient querying and long-term memory management, thereby improving the speed and accuracy of memory response.

    \item \textbf{Vectorization:}  Vectorized storage involves segmenting raw data into fragments and then converting them into high-dimensional vector forms, making similarity computation and rapid retrieval more convenient and efficient. Retrieval-Augmented Generation (RAG) holds an advantage in handling long-context tasks. By converting raw data into high-dimensional vectors, RAG significantly enhances the comprehensiveness and accuracy of retrieval and generation through vector matching techniques, fully leveraging its role in managing complex data tasks, thereby achieving efficient data processing.

    \item \textbf{Model parameterization:} Parameterized memory storage stores memory by adjusting model parameters (such as neural network weights) to realize information compression and generalization. Implementation methods include storing memory in the key-value cache of Transformer models or saving memory through parameter training in RNN-based models\cite{Sun-2024-Expressive}. Additionally, some methods involve fine-tuning large models to store memory.
\end{itemize}

These approaches transform raw data into different forms of long-term memory through various technical means, enhancing their adaptability to large models or intelligent agents in practical applications. Each method exhibits its own advantages and drawbacks in different scenarios. Currently, the most widely practiced approach is RAG-based storage. Meanwhile, the field of parameterized memory storage continues to see numerous emerging research outcomes. We have experimented and implemented multiple forms and plan to continue our research along this path.

\section{How can LTM be used to achieve model self-Evolution?}
\label{subsec:how_to_integrate}





After acquiring high-quality LTM data, the next challenge we face is how to use it to enhance model capabilities and enable model self-evolution. There are several key challenges need to be addressed in the process of using LTM data to maximize its effectiveness and efficiency, including:

\begin{itemize}
    \item \textbf{Adapting to Continuously Updated LTM Data}. As user LTM data is continuously accumulated, the model must balance learning new information while retaining previously acquired knowledge. Traditional models often assume stable data distributions, but in real scenarios, new LTM data may significantly diverge from earlier patterns, leading to risks like overfitting or catastrophic forgetting. Efficiently handling these shifts is critical to adapt to the dynamic LTM data.
    \item \textbf{Real-Time Learning and Efficient Feedback Integration}. Due to the LTM data being accumulated dynamically, models must quickly adapt to real-time changes in user behavior. Rapid integration of new data is vital for applications such as intelligent assistants, where seamless user interaction is key. Besides, user feedback, both implicit (e.g., clicks or time spent) and explicit, should be taken into consideration when refining the foundation model. 
    Incorporating both types of feedback in real-time enables the model to continuously improve and align with individual user needs.
    \item \textbf{Handling Data Sparsity and User Diversity}. Data sparsity is a common issue in continuously updated LTM systems, especially for users with limited interaction history or sporadic activity, making it difficult to train models. Furthermore, User diversity adds further complexity, requiring the model to adapt to individual patterns while still generalizing effectively across diverse user groups.
\end{itemize}

In the following subsections, we will review existing efforts that can be adopted to solve these challenges and summarize their weaknesses. Note that LLMs have become the most powerful AI models, the model self-evolution discussion will also focus on using LTM combined with LLMs. The discussion is divided into three key sections: first, LTM as an external knowledge repository, with retrieval techniques like Retrieval-Augmented Generation (RAG) to access information dynamically (in \ref{chap6:LTM_as_ext}); second, the direct training of models with LTM, embedding knowledge into the model itself (in \ref{chap6:LTM_as_int}); and third, a hybrid approach that combines external retrieval for precision with a fine-tuned model for improved efficiency in recalling relevant data (in \ref{chap6:LTM_as_mix}).

\subsection{Incorporating LTM as Outside Knowledge Bases}
\label{chap6:LTM_as_ext}

Since LTM is generated during interactions and may be refined with preprocessing, which is continuously updated, an intuitive strategy of utilization is to keep it independent from the foundation model, serving as an outside knowledge base to assist with reasoning and inference. While taking Transformer-based LLMs as an example, their structure inherently does not retain or recall any previous state or sequence, which limits the model's ability to carry LTM information but only short-term ``working memory'' for the current context window. 
So extra efforts should be made to address this limitation in utilizing LTM as outside knowledge bases for LLMs.

The most popular strategy in using an outside knowledge base for LLMs is Retrieval-Augmented Generation (RAG) with In-Context Learning (ICL), which can be adapted to utilize LTM data too. ICL mimics working memory, processing short-term information. Meanwhile, RAG, serving as an external database, acts as a module for LTM.

As a comparison, the LLM's ICL ability to plan and reason improves as model size and capabilities grow, approaching—if not exceeding—the capacity of the human prefrontal cortex in specific domains. Human working memory is limited to a span of around 7±2 items, whereas modern LLMs, with increasingly large context windows (ranging from 4K to 1M tokens or more), can hold significantly more information. RAG, as an external storage mechanism, provides the potential for storing and retrieving an unlimited amount of information, which seemingly addresses the limitation of human memory by offering three analogous stages: encoding, storage, and generation.

\subsubsection{Advances in Model Memory Encoding, Storage, and Retrieval}

In this section, we first review existing efforts made in memory encoding, storage, and retrieval with RAG and ICL, which are necessary modules to utilize LTM to achieve model self-evolution. Although there are several challenges in distinct steps, techniques like adaptive memory pruning, more efficient indexing systems, and context-aware retrieval algorithms could help mitigate some of these issues. Additionally, integrating reinforcement learning-based feedback mechanisms could enable more selective memory encoding, allowing LLMs to prioritize important data and improve long-term retrieval accuracy.

\begin{itemize}
    \item \textbf{Encoding:} One of the most critical shortcomings of current LLM architectures is the indiscriminate nature of data encoding. Unlike human memory, which selects relevant information for long-term retention based on salience and context, LLMs encode vast amounts of information without prioritization, leading to bloated and inefficient memory systems. Recent advancements in RAG-based systems have focused on introducing mechanisms for selective encoding, similar to human memory. Models like MuRAG \cite{murag2022} and RA-CM3 \cite{racm322}, for instance, are integrating multimodal data (images, text) with refined hierarchical indexing systems. This approach mirrors how humans encode experiences from different sensory inputs, allowing the system to distinguish between relevant and irrelevant information, ensuring that only the most contextually pertinent data is encoded into memory. Additionally, recent techniques such as UPRISE \cite{uprise2023} offer promising methods to enhance memory encoding by constructing a pool of context-aware prompts, selectively retrieving relevant templates for different tasks. This dynamic process parallels how humans draw on past experiences when encoding new information, improving the efficiency and selectivity of memory formation.

    \item \textbf{Consolidation:} While RAG architectures allow for unlimited data storage, they still struggle with consolidation, organization, and fragmentation of memories, which become more difficult to retrieve as the memory system grows. In contrast, human brains consolidate memories during rest periods, integrating and organizing information in a way that strengthens important memories while letting others fade. To address this, recent studies haved developed more structured storage systems, such as RAPTOR \cite{raptor2024} and MEMWALKER \cite{memwalker2023}, which take a step towards improving the hierarchical organization of stored information. These models introduce a tree-like structure that clusters and summarizes data recursively, creating a more structured memory space that allows for efficient retrieval of information across different levels of abstraction.
    
    In addition, emerging models like $\text{Memory}^3$ \cite{Memory³} propose integrating explicit memory modules that manage knowledge as key-value pairs in external non-volatile storage, emulating how the human brain efficiently consolidates and accesses memories based on relevance and frequency of recall. Techniques like compressive memory, seen in Beyond Retrieval: Embracing Compressive Memory in Real-World Long-Term Conversations \cite{chen2024compressimpressunleashingpotential}, explore how to compress and store interactions while retaining the core elements of past experiences, thus reducing redundancy and improving access to the most relevant parts of memory during retrieval.

    \item \textbf{Retrieval:} The retrieval process is where the largest divergence between human and LLM-based memory systems occurs. While humans rely on associative and context-driven recall, current LLM retrieval relies heavily on probabilistic token generation, often resulting in the retrieval of incomplete or irrelevant information. Recent work has shifted toward improving retrieval systems by introducing hybrid search techniques that blend traditional keyword-based search with more advanced semantic retrieval systems. For instance, Blended RAG \cite{blendedrag2024} and Hybrid Search \cite{hybridsearch2024} combine BM25 with dense vector embeddings to improve both the relevance and context-awareness of retrieved information. Moreover, methods like FiD emphasize the need for creating more sophisticated queries that better match the intent and structure of the memory database, drawing parallels to how humans refine their recall based on the specific context of a query.
    
    Besides, post-retrieval mechanisms such as re-ranking and self-reflection, as seen in models like Self-RAG \cite{asai2023selfrag} and Re2G\cite{glass2022re2gretrievererankgenerate} , aim to refine the retrieved information to align more closely with the user’s query. These systems not only filter out irrelevant content but also prioritize the most contextually appropriate responses, echoing how human memory retrieval is shaped by the relevance and importance of past experiences to the current situation.

\end{itemize}

To summarize, the RAG and ICL strategies do not require updating the foundation model’s parameters, allowing for efficient integration and updating of LTM data, which makes it possible to support real-time, efficient LTM management and utilization. The foundation model can evolve with distinct LTM data through this strategy. However, despite improvements in memory encoding and retrieval, there is still a fundamental disconnect between the deep, associative nature of human memory and the more surface-level pattern recognition performed by LLMs. New models like MemoCRS \cite{xi2024memocrsmemoryenhancedsequentialconversational}  and COMEDY \cite{chen2024compressimpressunleashingpotential}  represent initial attempts to bridge this gap by integrating deeper, long-term contextual memory into conversational AI, but much work remains in improving the depth and dynamism of memory representation.

\subsection{Incorporating LTM by Updating Model Parameters}
\label{chap6:LTM_as_int}
Another type of strategy for model self-evolution using LTM is through a parameterized approach, where the accumulated LTM data is transformed into the model's internal parameter memory through model training. The largest difference between this method and the LTM data utilization in RAG and ICL is that it directly modifies the parameters of the foundation model. As a result, it requires a larger data scale, higher implementation costs, and lower update efficiency compared to the former. However, its advantage lies in not needing an explicit LTM management mechanism design, and it can also be completed through single-step inference during the inference.

Regardless of a foundation model's architecture, the LLM training paradigm is typically classified into three phases: pre-training, instruction tuning, and alignment tuning \citep{zhao2023survey}. Each phase serves a distinct purpose and many state-of-the-art LLMs \citep{touvron2023llama2, dubey2024llama3, jiang2024mixtral} undergo multiple iterations (and often interleaving) of these phases to achieve optimal results. In our context, each phase presents a unique opportunity for LTM data integration.

\subsubsection{Brief Review of LLM training}

We first briefly review each of the LLM training phases and discuss how they can be utilized to encode LTM data for model self-evolution.

\textbf{Pre-training.} LLMs undergo the pre-training phase in an unsupervised manner and typically using diverse, large-scale corpora. This is essential for LLMs because it can equip them with extensive world knowledge and a deep understanding of language \citep{brown2020fslearner, zhao2023survey}. 
The next-token prediction task is commonly used to train decoder-only models, such as Llama \citep{touvron2023llama1} and GPT \citep{brown2020fslearner}, which aims to predict a target token $x_i$ given the preceding sequence in an auto-regressive manner. Concretely, we would like to maximize:
\begin{equation}
\mathcal{L} = \sum_{i=1}^n \log P(x_i | x_1,\ldots,x_{i-1}).
\end{equation}
Pre-training typically requires significant computational resources and large-scale data support, while LTM data often faces data sparsity issues. So a candidate method known as continued pre-training can be employed to encode LTM into models. By using a similar training objective, a pre-trained LLM can continue learn new data about specific tasks \citep{parmar2024reuse}, such as recent information \citep{qin2022elle, jang2022conttrainllm} or domain-specific knowledge \citep{colombo2023saulllm, ke2022conttrainllmdomain, Gururangan2020conttraindomain}. Thus, this strategy also works for encoding LTM data. 

Despite continued pre-training enabling LLMs to acquire new knowledge and is particularly useful for integrating LTM data, there is a potential risk of catastrophic forgetting~\citep{parmar2024reuse}. Research shows that appropriate learning rate management \citep{gupta2023rewarm, winata2023lradjust} and careful adjustment of data distribution \citep{ibrahim2024simplecontpretrainstrat} to mitigate distribution shift play a critical role in a successful continued pre-training process. Additionally, techniques which enable updating only part of the model's parameters can be helpful to minimize the impact on the original knowledge \citep{wu2024llamablockexpansion}.

\textbf{Instruction tuning.} Instruction tuning is usually performed after pre-training to further enhance their capabilities. In this phase LLMs are fine-tuned with instruction instances in a supervised manner. These instances typically consist of a task description ($T$), along with input ($X$) and desired output ($Y$) pairs. LTM data can also be reorganized in this format and be encoded into LLMs. For LLM-orientated training, the instructions (or prompts) ($I$) are usually constructed as $I = T + X$ \citep{lou2024sftsurvey}. Therefore the training objective of instruction tuning (in the context of LLM, this is also called Supervised Fine-Tuning (SFT) \citep{ouyang2022rlhf}) that can be formalised as:
\begin{equation}
P\left(Y_1,\ldots, Y_n | I_1,\ldots,I_m\right) = \prod^{n}_{j=1} P \left(Y_j | I_1,\ldots, I_m,Y_1,\ldots,Y_{j-1}\right),
\end{equation}
where $I = \left\{I_1,\ldots,I_m\right\}$ and $Y = \left\{Y_1,\ldots,Y_n\right\}$ are the instruction and output sequence respectively. In essence, the model is trained to predict each token in $Y$ given $I$ and $Y$ so far. A common loss function for this training objective is \citep{shi2024sftloss}:
\begin{equation}
\mathcal{L} = -\log P(Y_1, \ldots, Y_n | I_1,\ldots, I_m) = -\sum_{j=1}^n \log P(Y_j | I_1,\ldots, I_m, Y_1,\ldots, Y_{j-1}).
\end{equation}

For LTM data, this phase presents a prime opportunity for memory integration. Unlike pre-training, only a moderate amount of data (i.e. pairs of ($I$, $Y$)) is required in this phase to significantly influence the model, leading to general performance improvements \citep{zhao2023survey}. As a direct comparison, humans receive a vast amount of real-time events, and they filter these signals to distill valuable information. Only a small portion of this information (i.e. $I$) is stored in human's LTM through repetitive activation and memorization, which can be retrieved and output (i.e. $Y$) later for downstream tasks. To mimic this process, we must curate a refined set of instruction data and train the model iteratively on this smaller set. The quality of this refined information (or data) is crucial in influencing both human and LLM decision-making and performance. Moreover, instruction tuning can also enables LLMs to generalize to unseen tasks \citep{chung2024sftscale}. Additionally, like pre-training, domain-specific knowledge can also be infused into the model, turning it into a domain expert. These qualities are highly desirable for effective LTM data utilization, as the final system should be capable of performing tailored and complex actions with self-evolved model. Understanding what data accurately reflects LTM is crucial when incorporating them into LLMs. The answer varies depending on the application. For example, in an emotional support chatbot, LTM includes past sessions between users and the chatbot \cite{li2024hello}. In a medical assistant scenario, LTM involves patients' past medical records \citep{zhang2023memory}.


\textbf{Alignment tuning.} Despite extensive data cleaning, the large volume of pre-training data may still vary in quality, potentially leading to undesired responses \citep{wang2024comprehensive}. To address this, several techniques have been proposed and implemented after the model’s fine-tuning process. These techniques, collectively referred to as alignment tuning, aim to align the model with human preferences. RLHF \citep{ouyang2022rlhf} and RLAIF \citep{lee2023rlaif} are two popular alignment methods, each with its own use case and limitations. RLHF tunes a reward model that rates different outputs based on human feedback, which is then used to further fine-tune the LLM \citep{christiano2017deep}. In contrast, RLAIF directly links the LLM to a larger, more aligned model for learning. Due to the complexity and overall performance of RLHF, other methods such as Direct Preference Optimization (DPO) \citep{rafailov2024direct}, Knowledge Transfer Optimization (KTO) \citep{kto}, Optimized Reward Preference Optimization (ORPO) \citep{hong2024orpo}, and Simplified Preference Optimization (SimPO) \citep{meng2024simpo} are often employed in different scenarios for alignment tuning. However, although it is technically feasible to adjust model preferences for specific applications or even incorporate domain knowledge in reward models to impart biases \citep{nath2024rmecommerce, lin2024dogerm}, curating a high-quality preference dataset that aligns with the desired LTM remains challenging.

In summary, model parameterization occurs across all three phases of LLM training, which theoretically allows for LTM integration at any stage of model self-evolution. However, due to constraints such as data availability, training stability, and computational resources, we believe that supervised fine-tuning (SFT) remains the predominant technique for direct LTM data integration. Achieving true model self-evolution, however, may require a more sophisticated approach, particularly in obtaining high-quality feedback signals for effective self-guidance.

\subsubsection{Advantage of Model Parameterization}
\label{model_personalisation_sft}
Compared to the LTM integration strategies based on RAG and ICL, parameterized encoding method has significant advantages in terms of input format and inference efficiency. As most current LTM data are in the textual form due to its clearer interpretability, better context control, and ease of implementation \citep{zhang2024memoagentsurvey}, supporting textual LTM requires LLMs to have a sufficiently large context window to incorporate all necessary information during prompt construction. However, decoder-based LLMs are typically constrained by a fixed, limited context window. To address this, significant efforts have been made to extend the context window effectively \citep{chen2023pi, press2022alibi, ding2024longrope, peng2023yarn, zhu2024pose, chen2024longlora, xiong2023longllama} or to process information in parallel \citep{hao2022structprompt, ratner2023parallelwindow}. Despite these advancements, the information stored in LTM can still exceed the capacity of state-of-the-art context windows. Additionally, due to the architecture of transformers \citep{vaswani2017attention}, which serve as the foundation for most LLMs, computational cost increases quadratically with input length. This makes extremely long inputs less practical for some applications. A limited context window necessitates the use of a retriever to choose relevant information from a large, manually curated LTM resource, such as a database or vector store \citep{zhong2024memorybank}. This shifts the challenge from the LLM to the design of an effective retriever. Injecting LTM data directly into the model addresses this limitation, as it eliminates the need to include such information in the input prompt. This allows the LLM to focus on the task itself, with memory retrieval integrated into the inference process. This integration leverages the LLM's advanced language understanding and reasoning capabilities, avoiding dependence on a (usually) less capable external retriever.

If we choose to use training methods to encode LTM data into LLMs, a vital question is how to decide the data structures and types for training. As we known, in LLM-based agents, short-term memory manages contextual information, while long-term memory stores past experiences, reflections, and profiles. Profiles are arguably one of the most critical components of an agent, directly influencing its functions and interactions \citep{wang2024autoagentsurvey}. Typically, profiles contain basic details such as age, gender, and occupation \citep{park2023stanfordtown}, and may also include traits, interests, and behavioral patterns \citep{wang2023userbehavsim}. While it is possible to fit such descriptions into a prompt, this approach only portrays an average person playing a specific role, lacking the depth needed to reflect an individual’s rich history and experiences \citep{shao2023charllm}. To integrate such data into the model in a unified way, one can use SFT with specially curated datasets. In Character-LLM \citep{shao2023charllm}, training data is constructed not only from profiles but also from experiences and scenarios, enabling LLMs to display more human-like characteristics. Similar works \citep{zhou2023charglm, chen2023haruhi} also fine-tune models using character-specific data derived from past memories and experiences, proving effective in aligning models more closely with their intended personas.

In addition to in-character memory, one of the most crucial aspects of LTM is its associated domain knowledge. It has been well established that LLMs can be significantly enhanced in specific domains through fine-tuning, enabling smaller models to potentially surpass larger models. This is especially evident in domains that require extensive knowledge for complex reasoning, such as healthcare and finance \cite{zhao2023survey}. For instance, the Med-PaLM series \citep{singhal2024medpalm1, singhal2024medpalm2}, fine-tuned from the PaLM model \citep{Chowdhery2024palm}, became the first model to pass the U.S. Medical Licensing Examination (USMLE), demonstrating human-expert level proficiency in the medical domain. In legal services, Legal Artificial Intelligence (LegalAI) \citep{zhong2020legalai} has been continuously evolving and has reached new heights with the advent of LLMs. Fine-tuned models can now perform in-depth legal question-answering, simulating the role of a lawyer \citep{huang2023lawyerllama, cui2023chatlaw}, and efforts are underway to develop intelligent legal service systems \citep{yue2023disclawllm}. In finance, domain-specific fine-tuning has also shown significant promise, as seen in models like BloombergGPT \citep{wu2023bloomberggpt} and FinGPT \citep{yang2023fingpt}. Domain knowledge plays a vital role in model self-evolution, and these examples demonstrate the feasibility of integrating specialized knowledge, which elevates models beyond their general-purpose capabilities.

\begin{figure}
    \centering
    \includegraphics[width=1\linewidth]{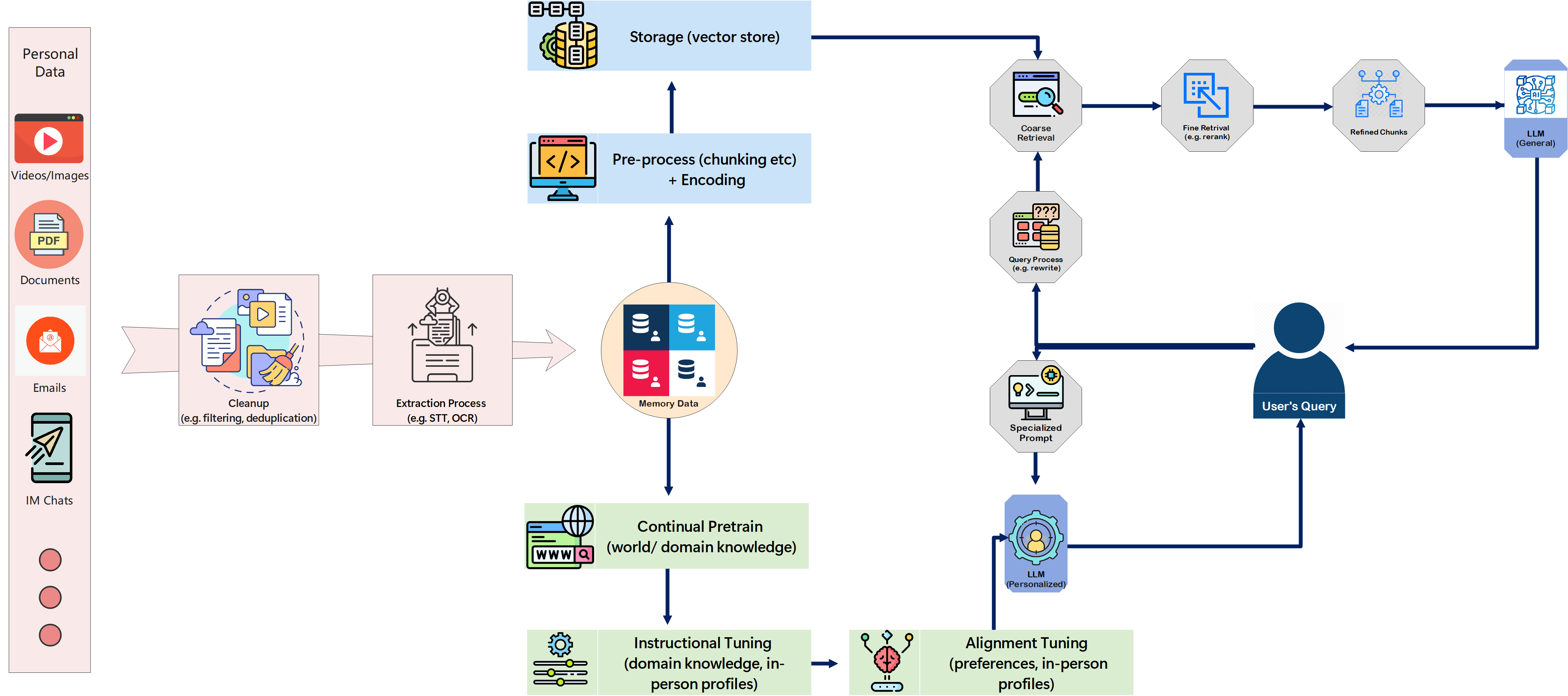}
    \caption{How personal data can be utilised through textual and parametric memory retrieval}
    \label{fig:memory_int_ext}
\end{figure}

\subsection{Incorporating LTM Data with Mixed Strategy}
\label{chap6:LTM_as_mix}
Since using LTM data solely as an external knowledge base or as additional training data each has its own advantages and limitations, a simple question is whether we can integrate these two methods (RAG with ICL and SFT) to leverage their strengths and achieve more effective model self-evolution. 
Recent works on RAG also see an increase in fine-tuning models at each stage of RAG: retrieval, augmentation, and generation. By focusing on fine-tuning, we can significantly enhance an RAG system’s efficiency and effectiveness in handling LTM data. 

\subsubsection{Fine-Tuning in the Retrieval Stage}
The retrieval stage in RAG systems involves selecting relevant documents or knowledge chunks from external sources to boost the model's intrinsic knowledge. Fine-tuning the retriever model at this stage optimizes its ability to retrieve more relevant, domain-specific information. 

By fine-tuning the retriever with LTM data, the precision of retrieving highly relevant content improves. In fields such as law or medicine, this ensures that the retriever pulls accurate and specialized content, increasing relevance and reducing noise during the retrieval process. This addresses the limitations of generic models, which may retrieve unrelated or less useful data \citep{gao2023ragllmsurvey}. This process is often applied to embedding models to align the representations of both the query and external documents with the domain's vocabulary and semantics. This alignment enhances retrieval effectiveness by improving semantic similarity matching. Moreover, fine-tuning has been used to align the retriever with the generator. One such method is the LM-Supervised Retriever (LSR), where the output of LLMs is used to train the retriever in a feedback loop \citep{gao2023ragllmsurvey}. For instance, REPLUG LSR \citep{shi2023replug} trains the retriever to select documents that can reduce the perplexity of the LLM's output sequences.

\subsubsection{Fine-Tuning in the Augmentation Stage}

The augmentation stage in RAG systems integrates retrieved LTM data with the model's existing knowledge to create a comprehensive context for generating an informed response. Fine-tuning can optimize the integration and selection of relevant information at this stage. Since the augmentation stage acts as an intermediary between retrieval and generation, fine-tuning can be applied to the retriever and/or the generator (or as an additional module in between).

For instance, fine-tuning enhances contextual filtering and selection by eliminating redundant or irrelevant information from the retrieved documents. A notable example is Self-RAG \citep{asai2023selfrag}, which uses reflection tokens to indicate whether retrieved documents are relevant to the query, whether the generated responses are supported by the documents, or whether the response effectively addresses the query. This approach significantly improves and streamlines the augmentation process. Additionally, models can optimize their augmentation strategies through document grounding, particularly when handling large volumes of documents. Fine-tuned models are better equipped to identify relevant and critical information from the retrieved content, resulting in more accurate response generation and reducing the likelihood of hallucination \citep{huang2020llmcite}.

\subsubsection{Fine-Tuning in the Generation Stage}

The generation stage is where LLM, enhanced by the retrieved context, produces the final response. Fine-tuning at this stage improves the quality of the generated text and ensures alignment with the desired style, tone, and factual accuracy. For instance, fine-tuning enables the LLM to conform to specific formats and enforce a particular response style. This technique is especially useful for equipping LLMs with function-calling capabilities and tool usage \citep{patil2023gorilla}.

Fine-tuning also allows the model to be customized for specific tasks or domains, ensuring that the generated content adheres to domain-specific norms and conventions. For example, in medical applications, fine-tuning helps the model generate clinically appropriate responses, while in legal tasks, it ensures adherence to legal terminology and reasoning. Additionally, the Chain-of-Thought process \citep{wei2022chain} can be integrated into model generation, guiding the model through complex reasoning and facilitating connections between relevant knowledge in the retrieved documents \citep{li2024rtghop}.

In summary, the integrated LTM utilization method combines the strengths of both strategies, optimizing data use and supporting model self-evolution. However, practical challenges remain, such as determining the optimal frequency of SFT to balance efficiency and effectiveness.

\subsection{Utilizing LTM with Multi-Agents}

Except for the primary encoding, storage, and retrieval studies about how to utilize LTM in specific LLMs, we propose that LTM can also be adopted by using RAG-based multi-agent systems to achieve model self-evolution. Several interconnected strategies can be employed to create a cohesive and efficient memory framework that mimics human cognition, including 1) dynamic memory storage, 2) context-based retrieval, 3) memory consolidation, and 4) self-reflection, all within a multi-agent framework. In this framework, different agents can understand LTM patterns of user behavior and improve their decision-making and action planning, and we will introduce them one by one. Note that each module can be implemented by a single LLM-powered agent.

\begin{enumerate}
    \item \textbf{Dynamic Memory Storage} is a key component in improving long-term memory is the dynamic management of stored information. Instead of passively accumulating data, memory systems must prioritize the retention of relevant information. This involves continuously updating the memory bank based on interaction context, user preferences, and relevance to the task. By applying a vectorized storage system, memories can be efficiently retrieved and updated, ensuring that the most pertinent information remains easily accessible. This approach mirrors the brain's ability to filter and prioritize experiences, focusing on valuable memories while allowing less critical information to fade.

    \item \textbf{Context-Based Retrieval} allows memory systems to dynamically recall relevant information based on current needs. Rather than relying solely on explicit keywords, retrieval can be enhanced by leveraging semantic similarity, temporal context, and retrieval frequency. This method improves both the precision and relevance of the information recalled, much like how humans retrieve memories by associating past experiences with present situations.

    \item \textbf{Memory consolidation} is also crucial for maintaining LTM for further utilization. Systems must periodically assess the relevance and usage of stored memories, reinforcing essential data through consolidation while gradually deprioritizing less frequently accessed information. This helps avoid issues such as catastrophic forgetting by ensuring critical memories remain intact while the system avoids becoming cluttered with outdated or irrelevant data.

    \item \textbf{Self-reflection} enables models to refine their LTM by periodically reviewing stored information. Through this reflective process, general principles can be abstracted from specific experiences, leading to more efficient memory management and retrieval strategies in the future. This self-reflective capacity not only optimizes memory storage but also enhances decision-making and adaptability over time.
    
\end{enumerate}

Comparing with previous LTM utilization strategies, the proposed multi-agent based LTM management framework are more general to corporate with distinct LLMs. The mentioned four modules/functions can be distributed across different agents with specific tasks. By analyzing user behavior patterns over time, agents can better plan actions and make more informed decisions, resulting in a more adaptive and personalized experience. The integration of these memory-focused processes allows agents to operate cohesively, leveraging their shared knowledge to create a more efficient and human-like system for managing and utilizing long-term memory. 

\subsection{Utilizing LTM to Support Agent Self-Evolution}

In previous subsections, we have reviewed the existing research and the multi-agent integration method we proposed to use RAG and ICL for model self-evolution with LTM. Furthermore, take the Agent Hospital study conducted by the Tsinghua University team as an example~\citep{li2024agent}, we will show how the use of LTM helps improve the model’s self-capabilities in a specific simulacrum medical scenario. The proposed LTM accumulated and utilization method MedAgent-Zero can be mainly divided into three modules: medical records accumulation, medical experience reflection, and RAG-based LTM utilization.

\begin{enumerate}
    \item \textbf{Medical Records Accumulation.} In this work, the doctor agent first accumulates LTM from successful medical cases to handle future problems. Taking disease diagnosis as an example, after learning the symptoms of a virtual patient, if the doctor agent provides a correct response, this diagnostic process will be saved as a medical record, just like a real-world doctor does, forming text-based LTM data. It is worth noting that even in the real world, the accumulation of medical records is very beneficial for enhancing a doctor's medical capabilities, as it can provide strong decision support when encountering similar issues in the future.

    \item \textbf{Medical Experience Reflection.} In the second module, we focus on how the doctor agent accumulates LTM through self-reflection from failed medical cases. Therefore, unlike storing successful medical cases in the memory base directly, we use self-reflection techniques in this module to allow the doctor agent to reflect and summarize after answering the diagnostic problem. The original question, its own response, and the correct response are adopted to generate experience-based LTM data. These data are also text-based, and to ensure their accuracy, we verify if the doctor agent can correctly answer the original question after utilizing the summarized experience (LTM will be dropped once cannot correctly answer). In other words, we designed the construction method for LTM data based on the agent's interaction with different outcomes.

    \item \textbf{RAG-based LTM Utilization.} Based on the two types of accumulated LTM data mentioned above, we propose using RAG technology to first retrieve stored cases and experiences when addressing new problems. It is important to note that the retrieval is based on the questions that generated the LTM data. Besides, since no parameter updates are required throughout the process, experience accumulation and application can be conducted online by ICL, making efficient use of the LTM data. As the doctor agent continually acquires valid LTM data during inference, its ability to handle medical questions also improves. Experimental results show that after self-evolving through diagnosing tens of thousands of patients, the doctor agent’s diagnostic capabilities in the respiratory department even surpassed the current SOTA model trained on real-world data.
\end {enumerate}

To summarize, similar to the accumulation and inheritance of human knowledge that can help the development of culture, with appropriate algorithmic support, LTM can also effectively help models in achieving self-capability enhancement, which is a vital step in model self-evolution. Although the handling and application types of LTM in this work still have limitations, it fully demonstrates the potential of this direction. In Section ~\ref{subsec:main_feature_plan}, we will provide future optimization directions.
\section{The Practice of model self-evolution based on LTM}
\label{subsec:our_practice}





\subsection{Practice on LTM Data Acquisition}
In our quest to enhance LTM data retention and accessibility, we have embarked on a comprehensive investigation of various methodologies. This exploration encompasses a spectrum of approaches, each designed to augment the robustness and reliability of LTM.
Firstly, we have delved into the realm of real-world data collection. This involves the systematic gathering of data from actual medical environments and scenarios, ensuring that the information is as authentic and representative as possible. 
Secondly, we have experimented with real synthetic data generation. This innovative approach \cite{murtaza2023synthetic} amalgamates actual data with artificially generated data to create a more comprehensive dataset. These synthetic data are crafted to fill in gaps where real data may be scarce or insufficient, thereby providing a more balanced and extensive pool of information for analysis and learning.
Lastly, we have ventured into fully synthetic data generation. This method relies entirely on artificially created data with techniques like Chain-of-Thought\cite{liang2023promptinglargelanguagemodels}, which can be tailored to meet specific requirements or simulate particular conditions that may be difficult to replicate in real-world scenarios. The flexibility of synthetic data allows for exploring a wide array of possibilities, pushing the boundaries of what can be achieved with LTM data.
Through these diverse tryouts, we strive to unlock the full potential of LTM data, paving the way for more efficient and effective model evolution.

\subsubsection{Real-world LTM Data Collection}
\label{chap7: data collection smhc1}
Developing robust data collection and labeling protocols is crucial for ensuring diversity and high quality in long-term memory (LTM) datasets, especially in specialized fields like healthcare and mental health. To meet this need, we collaborated with the Tianqiao and Chrissy Chen Institute (TCCI) and Shanghai Mental Health Center (SMHC) to create a sophisticated data collection framework. This framework is used in psychological consultations at two SMHC branches in Xuhui and Minhang districts, collecting real-world outpatient data between psychiatrists and incoming patients.

Our framework design prioritizes data diversity and quality. The collection process combines oral (recorded consultations) and written (consent forms, clinical notes) interactions, resulting in a comprehensive and heterogeneous dataset essential for building strong LTM. Adherence to standard operating procedures (SOPs) during patient interactions ensures consistency and quality across the dataset. Clinicians follow specific protocols, such as capturing vital signs and recording spontaneous speech, to maintain consistency across sessions. 

Ethical and privacy considerations represent foundational pillars of this process. Informed consent was meticulously secured, with nurses providing detailed explanations to ensure that patients understood the recording procedures. Written consents were securely tracked and managed. Post-collection, transcription services (e.g., iFlytek) were employed to convert audio data into text, followed by a rigorous anonymization process overseen by research assistants to safeguard privacy and ensure ethical compliance.

In addition, a well-structured operational framework with clearly delineated roles was implemented to enhance both accountability and efficiency. Nurses introduced the project and gathered consent, while research assistants supervised the process. Clinicians executed the SOP during consultations. A coordinated protocol for data management and device usage ensured timely collection, transfer, and calibration of both data and equipment, with responsibilities distributed between research assistants and physicians. This refined approach integrates diverse data sources while maintaining high-quality standards, which is crucial for collecting LTM data and developing robust AI systems in specialized fields like mental health.

Specifically, the LTM dataset collection encompassed over 1,000 participants from SMHC, accumulating more than 30,000 minutes of high-quality doctor-patient audio recordings. Participants ranged in age from 12 to 80 years old, were required to be fluent in Mandarin Chinese, capable of providing informed consent, and free from significant physical illnesses. Following all consultations, professional physicians conducted diagnoses using ICD-10 and DSM-5, meticulously documenting the present illness history, chief complaints, and prescribed medications, among other details. As previously mentioned, all recordings and associated metadata underwent careful anonymization to protect participant privacy.

From these recordings, we filtered out the participants who came for the follow-up consultation and verified 1,160 sample data points to create an LTM dataset in psychiatry (referred to as the SMHC dataset). The dataset comprises 553 participants diagnosed with Major Depressive Disorder (MDD), 426 diagnosed with Anxiety Disorder (AD), and 181 individuals classified as "Other" (neither MDD nor AD), with a total number of 25 different diagnoses. Diagnoses were made using the Chinese version of ICD-10. The demographic characteristics of the sample are presented in Table \ref{tab:demographics}.

\begin{table}[htbp]
  \centering
  \caption{Demographics of all participants in SMHC study.}
  \label{tab:demographics}
  \setlength\tabcolsep{2pt}
    \begin{tabular*}{\linewidth}{@{\extracolsep{\fill}}lccc@{}}
    \toprule
          & \multicolumn{1}{c}{\textbf{MDD (N = 553)}} & \multicolumn{1}{c}{\textbf{AD (N = 426)}} & \multicolumn{1}{c}{\textbf{Others (N = 181)}} \\
    \midrule
    \textbf{Age (years)} & 29.2 ± 9.95 & 34.04 ± 12.03 & 27.46 ± 11.11 \\
    \textbf{Gender (\%)} &       &       &  \\
    \textbf{- Male} & 176 (31.8) & 138 (32.4) & 77 (42.5) \\
    \textbf{- Female} & 377 (68.2) & 288 (67.6) & 104 (57.5) \\
    \bottomrule
    \end{tabular*}%
  \label{tab:addlabel}%
\end{table}%

The applicability of this thorough, ethically compliant, and technologically integrated workflow extends beyond health management. For instance, in diverse business contexts such as office collaboration, deploying similarly tailored data systems has shown equally promising results. This approach not only ensures data diversity and quality but also addresses ethical considerations and operational efficiency, ultimately laying a strong foundation for constructing personalized LTM data.

\subsubsection{Synthetic LTM Data Acquisition} 

\begin{figure*}[ht]
\centering
\includegraphics[width=\linewidth]{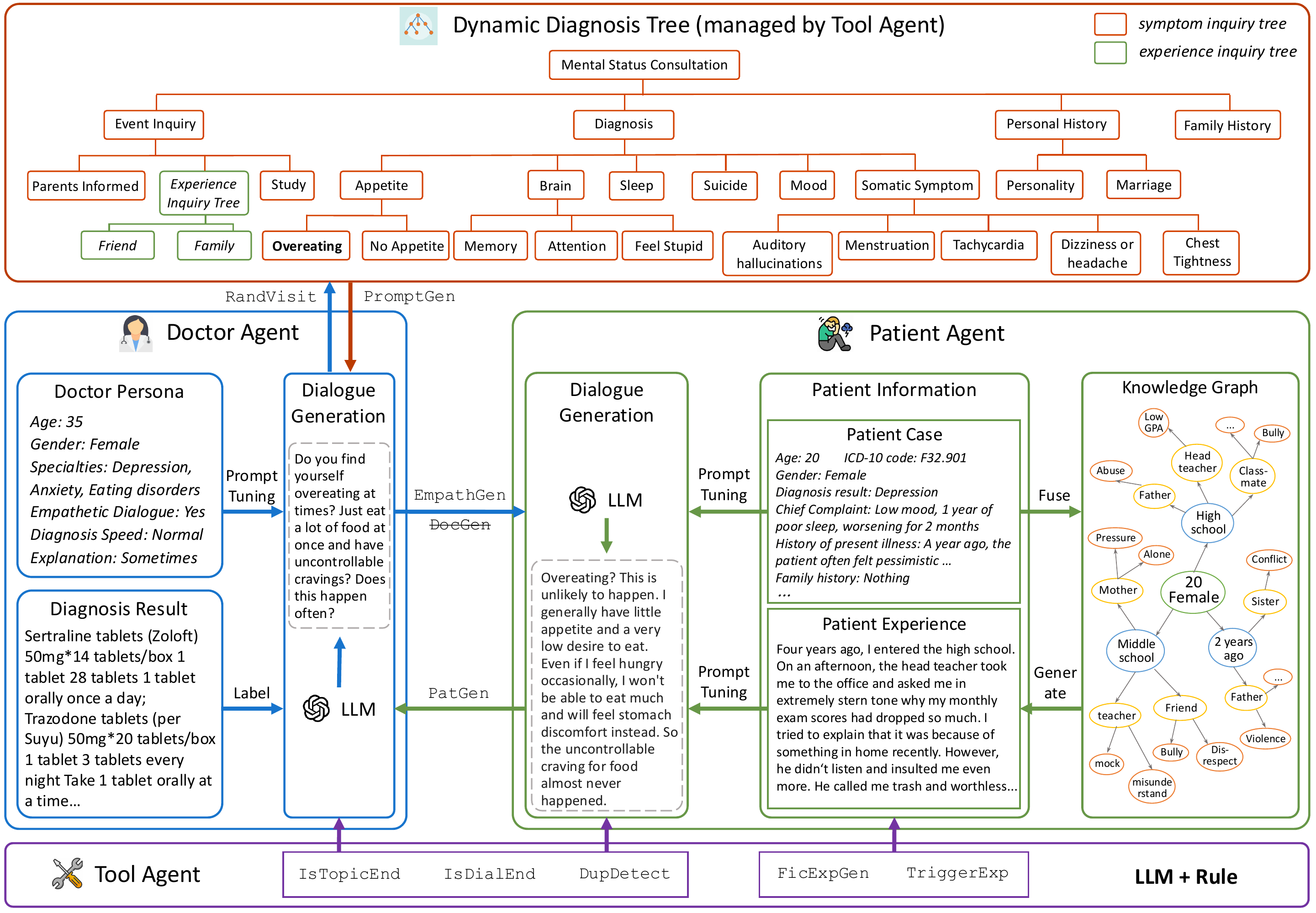}
\caption{The neuro-symbolic multi-agent framework for synthesizing diagnostic conversation of mental disorders.}
\label{mdd}
\end{figure*}
\subsubsection*{Real Data Enhanced Synthetic LTM Generation}
\label{chap7:data_synthetic_memtal}
While real-world data collection produces high-quality data, it requires significant time, effort, and other resources. To alleviate this challenge, real-data-enhanced synthetic data generation can be a viable option. Notably, the data collected as described in \label{chap7: data collection smhc} is promising to boost the AI mental healthcare community. However, it cannot be directly used publicly due to stringent privacy and ethical considerations. To address this issue, we propose synthesizing diagnostic conversations by utilizing anonymized patient cases collected in this study that are more accessible. We introduce a neuro-symbolic multi-agent framework that takes patient cases as input to generate synthetic diagnostic conversations of mental disorders. As illustrated in Figure \ref{mdd}, this framework involves three types of large language model agents: a doctor agent, a patient agent, and a symbolic tool agent responsible for managing diagnostic topic shifts. This framework features two major innovations: 

    (\romannumeral1) One-to-many patient case-to-dialogue generation that maximizes the utilization of precious real patient cases. Unlike previous studies \citep{zhang2024cpsycoun,wang2024notechatdatasetsyntheticdoctorpatient} that generate one conversation with one patient case. Our proposed framework is capable of generating multiple diverse diagnostic conversations with one single patient case. Specifically, three methods ensure the diversity and correctness of the diagnostic process. 
    
\begin{itemize}
    \item First, doctor agents with different diagnosis habits are designed and randomly selected for each conversation.
    \item Second, we use LLM with the knowledge graph to generate multiple fictitious patient experiences based on one patient case. The patient experiences serve as background information for patient agents during generation. Since the diagnosis of mental disorders mainly relies on symptoms rather than concrete events, integrating the fictitious patient experiences enhances the diversity of synthesized conversation while maintaining the accuracy of the diagnostic process.
    \item Third, the sequence of diagnostic topics is randomly determined for each conversation.
\end{itemize}

    (\romannumeral2) Another significant innovation lies in text generation under symbolic control via a dynamic diagnosis tree. This tree consists of a fixed symptom inquiry tree and a dynamic experience inquiry tree. Current clinical diagnosis of mental disorders strictly follows standards from ICD-11 \citep{world2018icd} or DSM-5 \citep{american2013diagnostic}. To simulate this process, we design the fixed symptom inquiry tree based on Structured Clinical Interview for DSM-5 (SCID-5) \citep{first2014structured}, covering all the diagnostic topics for important symptoms inquiry. The experience inquiry tree is constructed by extracting possible topics from the patient's response to past experiences. It's designed to establish deeper engagements with the patient. 

By applying the proposed framework and leveraging the SMHC real data, we further utilized LLM to generate synthesized data and release the largest Chinese \textbf{M}ental \textbf{D}isorder \textbf{D}iagnosis dataset MDD-5k. It is also the first labeled mental disorders diagnostic conversation dataset with diagnosis results from professional psychiatrists.

MDD-5k contains 5000 high-quality diagnostic conversations, averaging 26.8 turns and 6906.8 Chinese words per dialogue. It is built upon the SMHC dataset mentioned in Section \ref{chap7: data collection smhc1}, covering more than 25 different diseases (with a majority of MDD and AD). All cases have been cleaned and filtered by global standards to ensure the complete protection of patient private information. To reinforce the clinical setting, professional psychiatrists and psychotherapists supervise the whole process and filter out unqualified dialogues. In addition, they provide diagnosis summaries based on the history of the portrait and dialogue. As shown in Table \ref{mdd_result}, the synthesized MDD-5k dataset outperforms other diagnostic conversation datasets in human evaluation.
\begin{table*}
\centering
\resizebox{\textwidth}{!}
{
\begin{tabular}{cccccccc}
\toprule
\textbf{Dataset} & \textbf{Professionalism} & \textbf{Communication (\romannumeral1)} & \textbf{Communication (\romannumeral2)} & \textbf{Fluency (\romannumeral1)} & \textbf{Fluency (\romannumeral2)}& \textbf{Similarity} & \textbf{Safety}\\
\midrule
$\text{D}^4$ & 6.6 & 7.9 & 7.8 & \textbf{8.6} & \textbf{8.2} & 7.2 & \textbf{0}\\
CPsyCounD & 5.2 & 5.4 & 5.6 & 8.4 & 8.0 & 4.4 & \textbf{0}\\
Role-playing & 6.8 & 6.6 & 7.2 & 6.9 & 5.5 & 6.4 & \textbf{0} \\
MDD-5k & \textbf{8.6} & \textbf{8.3} &\textbf{8.4} &  \textbf{8.6}& 7.6 &\textbf{8.8}& \textbf{0}\\
\bottomrule
\end{tabular}
}
\caption{Human evaluation of different datasets.}
\label{mdd_result}
\end{table*}

\subsubsection*{COT Enhanced Synthetic LTM Generation}
\label{chap7.1.3: data generation RTG}
Although we can acquire LTM data by collecting personalized interaction histories and human feedback from real-world scenarios, this strategy is often limited by high collection costs and low efficiency, making it difficult to achieve large-scale data collection quickly. In particular, real-world data often lacks key steps in the reasoning process. Therefore, we propose using LLM generation to supplement the reasoning process, thereby constructing more comprehensive LTM data for the study of model self-evolution.

To address the issue of missing intermediate reasoning steps in the thought process, we introduce a technique called Retrieval-Thinking-Generation (RTG) for LTM data synthesis, designed to enhance memory recall. The core idea is to leverage Chain-of-Thought (CoT) reasoning to strengthen Retrieval-Augmented Generation (RAG) capabilities. 

The generated LTM data should contribute to training model, especially for RAG-based strategies. Therefore, the generated data needs to include not only the LTM positive data that should be utilized but also negative samples (distractors) that should not be used for RAG models. The key to constructing helpful training data is to construct high-quality distractors, so our designed RTG complete process is as follows, and an example of an RTG prompt is provided in Appendix \ref{appendix_A}. 

\begin{itemize}
    \item First, we can select high-quality question-answer pairs from existing datasets as the foundational model input. These sources can include Wikipedia-based Q\&A, web searches, or domain-specific datasets.
    \item Second, we focus on constructing high-quality distractors for the existing Q\&A pairs. These two types of data should have a high degree of similarity (to train the model’s discriminative capability), but the distractors should not be able to answer the question. We can complete the selection process by using a context similarity-based strategy to filter related paragraphs from the original corpus used to construct the Q\&A pairs and then verify them.
    \item Finally, based on the Q\&A pairs along with distractors (which can range from several to dozens), we will use CoT (Chain of Thought) Reasoning to divide the data utilization into several explicit reasoning processes, including: 1) Determining whether the current paragraph is related to the question; 2) If related, does it help answer the question? 3) Combining all paragraphs that can help answer the question to generate a response and list all cited documents. Such a reasoning and utilization process will help improve the use details of LTM data.
\end{itemize}

In addition, since LLMs may generate hallucinations, we ensure the constructed data meets correctness through the following validations: 1) The index of the ground-truth paragraph should be cited; 2) The paragraphs marked as cited should appear in the citation list; 3) All cited documents should be in the given candidate set.

To verify the effectiveness of the RTG strategy, we constructed training data for LTM utilization based on this approach and annotated a small-scale test dataset to evaluate its performance.

\textbf{Training Data Generation} 

To validate LLMs's memory retrieval capability, we constructed a custom training dataset based on our RTG method. The two primary sources for this dataset are the SQuADv2 dataset \cite{rajpurkar2018know} and WebGLM\cite{liu2023webglm}, which focus on Wikipedia and web data, respectively. A key distinction between our dataset and the original datasets is the introduction of a curated list of distractors, built on top of the provided context or references.

For Wikipedia data, we calculated embeddings of the context and retrieved the most similar paragraphs from Wikipedia\footnote{Cohere Multilingual Embeddings for Wikipedia in 300+ Languages: \url{ https://huggingface.co/datasets/Cohere/wikipedia-2023-11-embed-multilingual-v3}} to generate the distractors. For web-based data, we used SerpApi\footnote{https://serpapi.com/} to search for relevant questions and built distractors from the most similar content chunks. Notably, we applied the Levenshtein distance to filter out paragraphs or chunks that are highly similar to the original context of reference. We utilized the GPT-4o API to perform reasoning and generate results with correct citations.

During data construction, we observed that not all questions in the original dataset are self-contained, meaning additional context is required for the question to be fully understood. For example, in SQuADv2, the question "In what R\&B group was she the lead singer?" is ambiguous, as "she" should refer to Beyonce. Such ambiguity can potentially lead to the generation of random distractors, thereby lowering the overall quality of the dataset. To address this issue, techniques such as coreference resolution \citep{clark2016neuralcoref} were employed to filter out incomplete questions by identifying unresolved coreferences.

Finally, we curated a training dataset comprising 40,000 data instances. Each instance contains a question, 20 passages (including an average of 1-3 gold passages), the document ID of the relevant passage, a CoT process with citations and quotes, and the answers. The ratio of Wikipedia to web-based data is approximately equal.


\textbf{Evaluation Dataset Construction}

To address the limitations of existing Q\&A evaluation datasets, where candidate documents vary in length, lack targeted design for distractors, and use single-dimensional metrics, making them unsuitable for evaluating the application capabilities of LTM data after training with the generated dataset, we have constructed a dataset called LTM-COT-1. 

LTM-COT-1 consists of 158 instances with high-quality preprocessing and annotations. Each instance is composed of two parts: (1) a question-answer pair with corresponding contexts and supporting evidence, and (2) 19 distractor passages. The constructing strategy is similar to the training set.

The dataset is derived from the SQuADv2 validation set \cite{rajpurkar2018know}, where five annotators provided answers to the questions. To minimize potential ambiguity, we selected only those instances where the annotators' responses were fully consistent. To address edge cases where distractor passages might inadvertently contain the correct answer, we employed the APIs from OpenAI\footnote{https://openai.com/} and Anthropic\footnote{https://claude.ai/} to conduct verification. Finally, five human annotators conducted an additional review to confirm data quality and eliminate any inconsistencies.

\subsection{Practice on LTM Data Utilization}

\subsubsection{LTM Utilization with SFT and RAG}
\label{subsec:sft_utl}
We used the private LTM-COT-1 dataset along with several publicly available RAG-related datasets to test whether the model demonstrates enhanced data utilization capability after training with the constructed LTM data. We selected the Llama-3-70B-instruct version as the backbone model and performed SFT training using the generated LTM data, the trained model is named as Homer-70B.

\subsubsection*{Experimental results on LTM-COT-1}
We use two metrics to evaluate model performance here: 1) Answer Accuracy, calculated by dividing the number of correct answers by the total number of questions, where GPT-4 is adopted to judge the correctness. 2) Cite Score, indicating the performance of selecting citation paragraphs in terms of both accuracy and recall. Each correct earns one point and incorrect reduces one point. The final score is obtained by dividing the number of correct citations. 

In Table \ref{tab:benchmark_results}, we present the evaluation results of various prominent LLMs on our private benchmark dataset, and the prompt settings are also listed in the table. We can see that: 1) After training with the proposed synthetic dataset, our Homer-70B achieved the best in both Answer Accuracy and Citation Score, showing that reasoning data generation is helpful in achieving better utilization. 2) GPT-3.5 and Command R+ often cite multiple documents to avoid missing the golden document, but only get an undesirable behavior. GPT-4o, using vanilla RAG instead of RTG prompts to align with current industry settings, performs the second showing its strong ability. 3) The base model Llama3-70B has a weak ability to select paragraphs, also showing the usefulness of the generated dataset.

\begin{table}[htb]
\centering
\caption{Experimental results on LTM-COT-1 benchmark}
\label{tab:benchmark_results}
\begin{threeparttable}[b]
\begin{tabular}{l|ll}
\hline
Model                                                 & Answer Accuracy              & Citation Score           \\
\hline
GPT-3.5\tnote{1}                                 & 91.8\% & 56.3\% \\
Command R+ \tnote{2}                       & 96.8\% & 54.1\%  \\
Llama3-70B-instruct\tnote{3} & 97.4\% & 76.2\% \\
GPT-4o\tnote{1}                                 & 98.1\% & 86.7\% \\
\textbf{Homer-70B}                               & \textbf{98.7\%} & \textbf{91.2\%}  \\ \hline
\end{tabular}
   \begin{tablenotes}
     \item[1] Vanilla RAG.
     \item[2] Official API.
     \item[3] RTG prompts w/o training.
   \end{tablenotes}
\end{threeparttable}
\end{table}

\subsubsection*{Experimental results on public benchmarks.} 
Furthermore, we evaluated our model on a diverse set of standard academic benchmarks: ARC, HellaSwag, MMLU, TruthfulQA, Winogrande, and GSM8K in Table \ref{tab:academic_results}. 
The results show that our model achieves the highest average score across all compared models,  indicating that despite being tested on various tasks, our model retains strong generalization performance learned from the generated RTG dataset.

\begin{table}[htb]
\centering
\caption{Experimental results on distinct public benchmarks.}
\label{tab:academic_results}
\begin{tabular}{lllll}
\hline
Model      & DBRX Instruct & CommandR Plus & Llama3-70B-instruct & Homer-70B \\
\hline
Model Size & A36B/132B     & 104B                       & 70B                      & 70B       \\
\hline
ARC        & 68.90\%         & 70.99\%                      & 71.42\%                    & \textbf{72.70\%}     \\
HellaSwag  & 89.00\%         & \textbf{88.60\%}                      & 85.69\%                    & 88.13\%     \\
MMLU       & 73.70\%         & 75.70\%                      & \textbf{80.06\%}                    & 79.81\%     \\
TruthfulQA & \textbf{66.90\%}         & 56.30\%                      & 61.81\%                    & 64.46\%     \\
Winogrande & \textbf{81.80\%}         & 85.40\%                      & 82.87\%                    & 81.61\%     \\
GSM8K      & 66.90\%         & 70.70\%                      & \textbf{85.44\%}                    & 83.78\%     \\ \hline
\textbf{Averaged}    & 74.50\%         & 74.60\%                      & 77.88\%                    & \textbf{78.42\%}     \\
\textbf{Delta}    & {-3.91\%}         & {-3.82\%}                       & {-0.53\%}                    &      \\
\hline
\end{tabular}
\end{table}

\subsubsection{LTM Utilization for Agent Self-Evolution in Medical Domains}

Currently, some studies have proposed distinct strategies to support agent self-evolution, but they mainly rely on the storage of interaction raw data and limited self-reflection. We believe that if the raw data can be preprocessed based on the concept of LTM, even when using the same evolution strategies, better performance should be achievable. Therefore, we adopted the MedAgent-Zero method from Tsinghua University's study~\cite{li2024agent}, but combined it with the LTM concept for improvement, and conducted the following experiment.

We first briefly review the core strategy of the MedAgent-Zero algorithm. By allowing the medical agent to interact with generated patients during diagnosis and treatment, the agent accumulates medical cases from correct interactions and reflects on experiences from incorrect interactions. After such evolution, its performance is evaluated on the respiratory subset of the MedQA dataset. Here, we focus on how to further improve and optimize the experiences gained from incorrect interactions, which is helpful in achieving better medical performance. Based on the idea of LTM construction, we believe that the experiences gained from single-interaction self-reflection should be further optimized to achieve better utilization.

Inspired by the necessary preprocessing of LTM, we propose that the experiences gained can be further refined through a few-shot prompting approach, making them more aligned with the task's specific needs and characteristics. In other words, raw experience will be rewritten into task-specific LTM. As part of the specific strategy, we first selected three high-quality LTM data samples through manual annotation and optimization in a preliminary experiment. The three LTM samples were then used as few-shot examples to rewrite all experiences gained during interactions, resulting in a new version of LTM experiences.


\begin{table}[]
\centering
\caption{Experimental results of MedAgent-Zero and LTM-enhanced MedAgent-Zero in the MedQA subset~\cite{li2024agent}. The best performance is in bold and \% is omitted for all results.}
\label{LTM-rewrite}
\resizebox{\linewidth}{!}{ 
\begin{tabular}{l*{10}{c}}
\toprule
\#Interactions    & 5,000 & 10,000 & 15,000 & 20,000 & 25,000 & 30,000 & 35,000 & 40,000 & 45,000 & 50,000 \\ 
\midrule
Original     & 93.06 & 94.44  & 93.06  & 94.44  & 91.67  & 94.44  & 93.06  & 93.06  & 91.67  & 91.67  \\ 
LTM-enhanced & 91.67 & 94.44  & 91.67  & 93.06  & 93.06  & 91.67  & 93.06  & \textbf{95.83}  & 94.44  & 93.06  \\
\bottomrule
\end{tabular}}
\end{table}

For the evaluation, we followed the existing work's approach by testing on the MedQA respiratory subset~\cite{li2024agent} and expanded the training scale to include 50,000 virtual patients. Experimental results are shown in Table~5, where we used a voting mechanism based on three repeated experiments to enhance stability. From the table, we can see that the LTM-rewritten strategy achieves better continuous evolution, avoiding the significant performance decline trend that can occur during interactions. While the original version saw a faster improvement in the initial phase, the unprocessed raw experience often led to conflicts as the number of interactions increased, eventually causing a decline in the model's performance. In contrast, the LTM-rewritten model achieved the best results in this subset (95.83\% accuracy), demonstrating the effectiveness of LTM in enhancing performance.

Although we only used a simple LTM rewriting method for this experiment, the results validated its effectiveness. We also believe that the introduction of LTM strategies will provide even greater benefits for self-evolution in broader scenarios.

\subsubsection{LTM Utilization with memory system design}

LLMs and LLM-powered agents demonstrate advanced language comprehension and can engage in high-quality interactions, making them promising tools for mental health diagnostics, particularly for depression~\cite{li2024agent}. However, several challenges are encountered in real-world application, including insufficient domain-specific knowledge, limited empathetic abilities, and an inability to learn and adapt based on prior experiences. To address these issues, the integration of LTM is crucial. 
\begin{figure}[hbtp]
    \centering
    \includegraphics[width=0.6\linewidth]{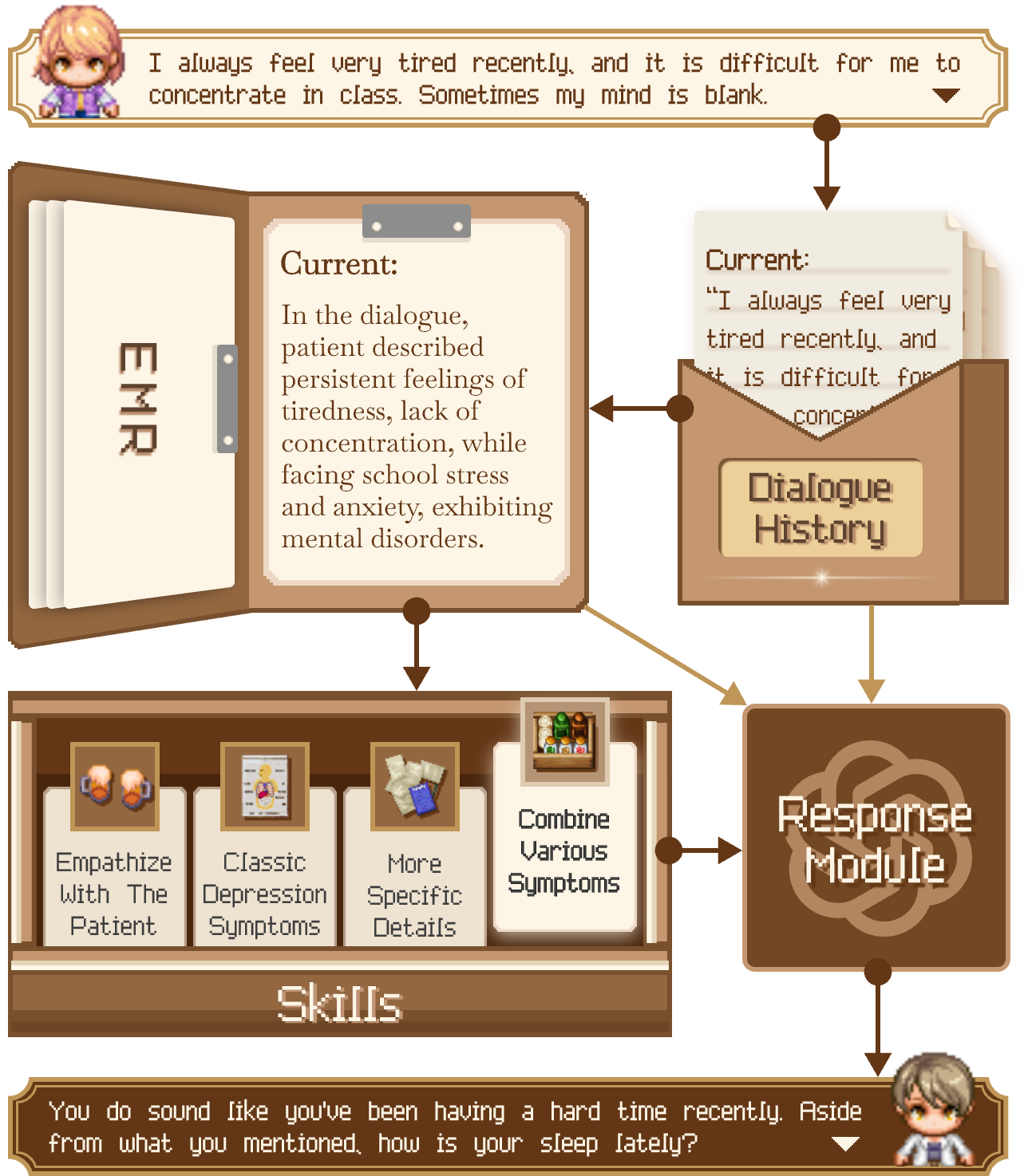}
    \caption{\textbf{The Tertiary Memory Structure.} The utterance of the diagnosis conversation will be stored in \textbf{Dialogue History}. The whole dialogue history in the session will be summarized into electronic medical records (\textbf{EMR}). \textbf{Skills} are generated by the supervisor plugin. All the memory will contribute to the dialogue generation.}
    \label{fig:memory structure}
\end{figure}
Inspired by the training process of psychiatrists, we propose an innovative three-level hierarchical LTM structure—\textit{Conversation Records}, \textit{Electronic Medical Records (EMR)}, and \textit{Diagnostic Skills}. This architecture enhances agents' ability to accumulate and apply knowledge, improving diagnostic accuracy. To evaluate its efficacy, we developed a novel conversational agent simulation system to mimic diagnostic sessions between patient agents and psychiatrist agents. This system generates realistic depression diagnostic conversations, providing a valuable tool for training intern psychiatrists and conducting preliminary depression risk assessments in individuals exhibiting depressive symptoms. Our experimental results, based on real-life depression diagnosis conversations D$^4$ dataset~\cite{Yao2022D4}, demonstrate an average improvement of 6.05\% in depression diagnosis accuracy and 1.8\% in suicide risk prediction. 

\subsubsection*{Tertiary LTM Mechanism}
The memory structure reflects the process psychiatrists follow: patient information is recorded through conversation, summarized into EMR, and used to refine diagnostic skills.  

\begin{itemize}
    \item \textbf{Conversation Records} represent the transcripts of the diagnosis conversation between patient agents and psychiatrist agents. It helps the psychiatrist agent to conclude the electronic medical records of the patient agents. 
    \item \textbf{Electronic Medical Records (EMR)} summarize patient information, including demographics, chief complaints, and symptomatology. They serve as a concise reference for diagnosing similar cases and for improving diagnostic skills. 
    \item\textbf{Diagnostic Skills} act as the optimizer in training psychiatrist agents. Diagnostic skills are generated by comparing the psychiatrist agent's diagnoses to ground truth diagnoses provided by professional psychiatrists in real life in D$^4$. These skills help refine the agents' diagnostic accuracy, guiding them to align more closely with professional psychiatrists by highlighting overestimated and underestimated diagnosed symptoms. 
\end{itemize}

As shown in Figure~\ref{fig:memory structure}, this three-tiered memory structure facilitates the abstraction and refinement of raw diagnostic information. Each layer serves as a distillation process, transforming detailed data into increasingly concise and enduring representations. This structure allows the system to store large amounts of data efficiently while ensuring that the most critical information is preserved for future retrieval, thereby enhancing diagnostic performance over time~\cite{lan2024depression}.

\begin{figure}[htbp]
    \centering
    \includegraphics[width=1\linewidth]{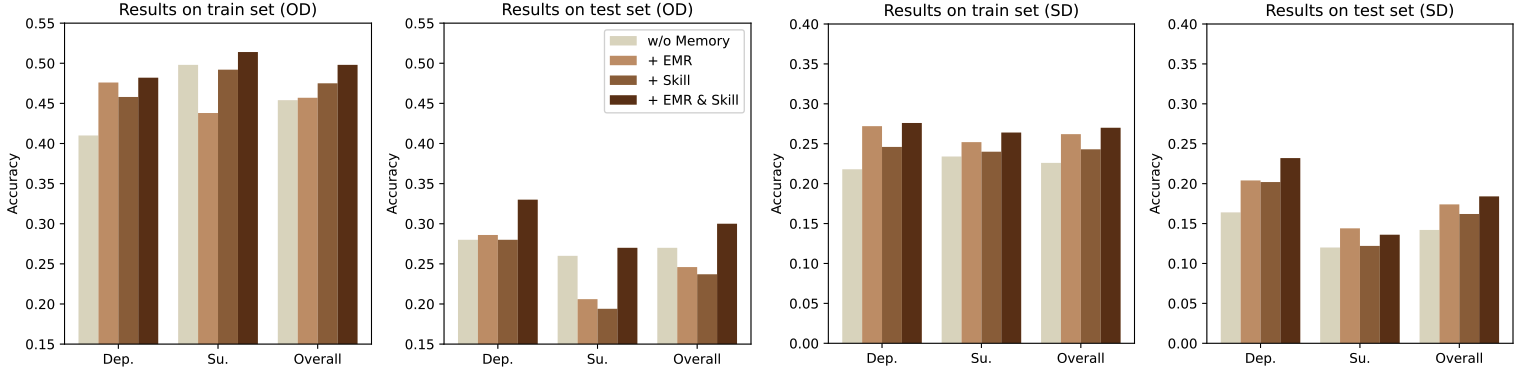}
    \caption{\textbf{The Results of Ablation Study on Memory Layers.} The first 2 pictures indicates the results(depression diagnosis and suicide risk prediction) based on the original dialogue history setting(OD), while the last 2 pictures implies the results based on the simulated dialogue setting(SD).}
    \label{fig:exp}
\end{figure}

We conducted series of experiment to assess the impact of these memory types on depression diagnosis by manipulating the memory module's activation: 1) No memory module, 2) Memory retrieval from EMR, 3) Memory retrieval based on diagnostic skills, and 4) Memory retrieval from both EMR and diagnostic skills. The results are depicted in Figure~\ref{fig:exp}.

The findings suggest that diagnostic skills are particularly effective in Original Dialogue (OD) settings, likely due to their alignment capabilities with LLM models. In Simulated Dialogue (SD) settings, EMR proves more beneficial, as the simulations often lack the precision of real-world conversations, and EMR can assist by referencing similar patient cases. Combining EMR and diagnostic skills yielded the most consistent and accurate results, outperforming LLMs without memory integration~\cite{lan2024depression}.

\subsubsection*{Effects of supervisor agent}

Another key innovation in this study is the introduction of a supervisor agent, which monitors the patient's symptoms, updates the symptom list, and generates targeted questions to assist the psychiatrist agent, streamlining the diagnostic process. It also manages the dialogue progression across predefined stages, improving diagnostic efficiency. This approach prevents the psychiatrist agent from asking redundant questions, instead guiding the conversation to focus on the unknown symptoms, thereby enhancing the efficiency and effectiveness of the diagnostic dialogue. Additionally, the supervisor agent manages the progression of the dialogue through three predefined stages of the diagnostic process~\cite{lan2024depression}.

We conducted an ablation experiment by disabling the reflection and question-generation, illustrating the fact that the supervisor agent significantly enhances risk prediction accuracy when its reflection and question-generation features are active.
Such a supervisory agent could be replaced by professional psychiatrists to monitor the data quality of both the doctor and patient agents. When the goal is to assist real patients in obtaining easier diagnoses, the efficiency and accuracy can be improved. When the aim is to train actual doctors using patient agents, the realism of the agents can be enhanced. When both the doctor and patient are agents, this can serve as a platform for generating simulated data under the supervision of real doctors.


\subsubsection{LTM Utilization with Real-time Weight Update}
\label{chap7:new_arch}
As we explore model self-evolution and the pivotal role of Long-Term Memory (LTM) in enhancing AI's cognitive abilities, it is crucial to explore innovations in neural architectures that could better facilitate these advancements. While Transformer architectures excel in reasoning and knowledge compression—typically through methods like supervised fine-tuning (SFT)—this knowledge is stored in the model’s weights, which are usually frozen during inference. As a result, real-time weight updates are not feasible in standard Transformer models, limiting the model’s ability to learn and adapt dynamically during inference. To overcome this, exploring new architectures, potentially integrating sequence-based structures like RNNs, could enable breakthroughs in LTM utilization and support more effective model self-evolution.

One notable study, \cite{Sun-2024-Expressive}, proposes a new class of sequence modeling layers known as Test-Time Training (TTT) layers. These layers are designed to overcome the limitations of existing RNN layers by transforming the hidden state into a machine learning model itself, updated through self-supervised learning steps. This innovative approach allows the model to continue learning even during test time, enhancing its performance on long-context tasks where traditional RNNs usually struggle. Thus, we propose to adopt TTT as the backbone structure to verify its performance in making use of LTM data, so some primary experiments are designed and conducted here.

\subsubsection*{Experimental Settings}

At the beginning of Section \ref{sec:issue_data}, we mentioned that the first challenge in using LTM data is adapting to continuously updated LTM data. So our experiment aims to verify the adaptability of TTT under distinct data distribution.

We choose multilingual datasets to conduct these experiments as distinct languages have various token distributions, where French, Chinese, and English are adopted.
By performing inference and updating model weights in different languages, we assess two questions: 1) whether the model can effectively learn new distribution patterns and improve its performance on the corresponding test sets. 2) whether TTT leads to catastrophic forgetting, i.e., whether the model's performance on the original test dataset deteriorates after adapting to the new distribution.

Specifically, we adopted the Book3 dataset \cite{gao2020pile} (a subset of The Pile) as the training dataset and trained a 1.3B TTT-linear model, denoted as $Model_{En}$. The Book3 dataset comprises a vast collection of English digitized books that range from classical literature to contemporary works, which has been widely used to train LLMs in long context. We selected two books for our experiments: "LE RÊVE DE SUZY" \cite{ardel1950rêve} as the French book and "The Smiling, Proud Wanderer (Xiào ào jiānghú)" \cite{jin1993smiling} as the Chinese book. After tokenization, we extracted 32,000 tokens from each book, with a fixed 2,000 tokens reserved as the test set. The remaining tokens were used as the training set, and we evaluated the impact of different training set lengths on model inference performance.


First, we performed inference on Model$_{En}$ using the training set and retained the inner-loop weights $W_{train}$ updated after inference. In this context, the process of parametric learning can be viewed as compressing a large amount of training data into the model’s weights, where the model effectively learns and stores underlying patterns from its training data. Specifically, each step of inference updates the weights $W_t$ by applying gradient descent on a self-supervised loss $\ell$. This process is akin to dynamically compressing the training data into a hidden state, which is then used to predict the next token. The update rule for the weights is defined as:
\begin{equation}
W_t = W_{t-1} - \eta \nabla \ell(W_{t-1}; x_t),
\end{equation}
where $\eta$ is the learning rate and $\ell$ is the self-supervised loss, which measures the difference between the model’s predicted token and the true token. The gradient of the loss with respect to $W$ drives the weight updates, enabling the model to adapt to new data distributions. This continual update mechanism ensures that the model captures significant input signals that produce large gradients, optimizing the weights accordingly.

Subsequently, we replaced Model$_{En}$'s initial weights with $W_{train}$, the updated weights after inference, resulting in the domain-specific models $Model'_{FR}$ and $Model'_{CN}$. These models are tailored to the respective French and Chinese datasets, having undergone inference and subsequent inner-loop weight updates on these new language-specific training sets.


Two experiments are designed to answer the two questions: 1) Experiment 1 is to compare the performance of Model$_{En}$ and Model$^{'}_{FR}$ / Model$^{'}_{CN}$ on the test set, where the training loss and perplexity (PPL) are adopted as metrics. 2) Experiment 2 tests the performance of Model$_{En}$ and Model$^{'}_{FR}$ / Model$^{'}_{CN}$ on the original Book3 test set to assess whether catastrophic forgetting occurs. To comprehensively evaluate the effect of TTT, we tried different lengths of train tokens and observed their impact on the performance of both the test tokens split from the Chinese/French books and the test tokens from the original Book3 pre-trained data.

\subsubsection*{Experimental Results}


\begin{figure}[h]
    \centering
    \includegraphics[width=1\linewidth]{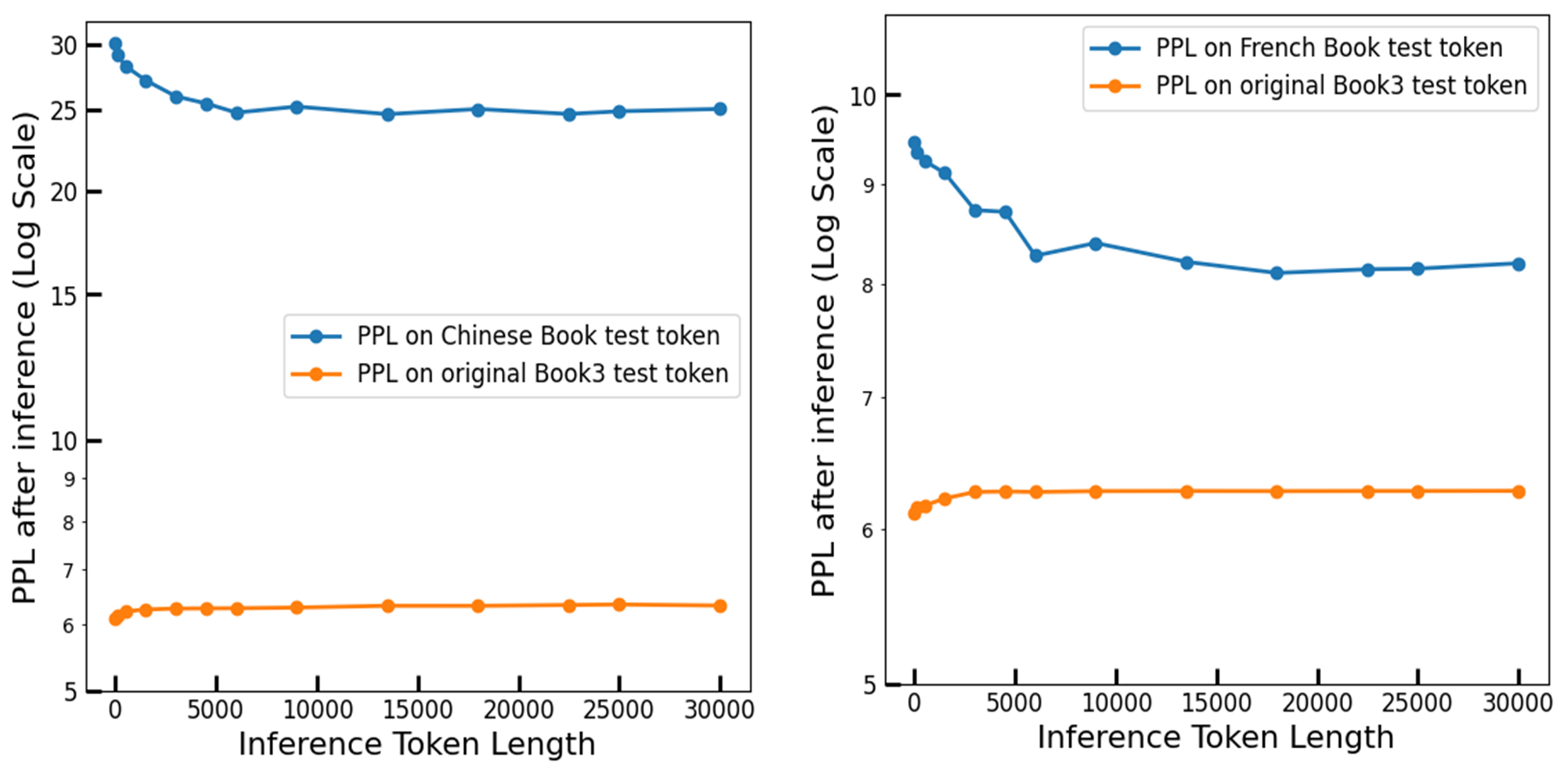}
\caption{\textbf{Left: PPL on test tokens after inference on Chinese books. Right: PPL on test tokens after inference on French books.} Both figures show the experiment results on learning new distributions and catastrophic forgetting tests. The figures illustrate the differences in perplexity (PPL) between using the initial weights $W_0$ \textit{(inference token length = 0)} and the updated weights $W_{{train}}$ after inference on Chinese and French books, measured on the corresponding test tokens. The results highlight how the model's performance shifts after inference on each language tokens.}
    \label{fig:ppl_fig_1}
\end{figure}

\textbf{Experiments 1: Learn on new distribution pattern.}

This experiment aimed to assess whether the TTT mechanism could effectively adapt to new data distributions and enhance model performance during inference.

As illustrated in Figure \ref{fig:ppl_fig_1}, the results demonstrate that by updating the model's weights ($W_0 \rightarrow W_{\text{train}}$) during inference on both the French and Chinese datasets, the model successfully adapted to the new distribution patterns. This adaptation led to a significant reduction in perplexity on the corresponding Chinese and French book test tokens. The plots show that as the inference token length increased, model performance improved consistently before stabilizing beyond a certain point. These findings confirm that TTT effectively learns from new language distributions, improving performance on both the French and Chinese test datasets, which directly addresses our first research question above.

\textbf{Experiment 2: Catastrophic Forgetting Test.}


The second experiment aimed to determine whether updating model weights through TTT inference leads to catastrophic forgetting, which would manifest as a significant performance drop on the original test set.

After conducting inference on the French and Chinese datasets and updating the weights, the model exhibited a minor increase in loss and perplexity when tested on the original Book3 dataset in English. The performance degradation was more pronounced after French inference compared to Chinese inference, though the impact was relatively limited overall, especially when shorter train token lengths were used. These findings suggest that while TTT does introduce some degree of catastrophic forgetting, the effect is minor, and the model largely retains its ability to generalize to the original data distribution.

To summarize, the experimental results demonstrate that TTT can effectively help the model learn new language distributions and improve performance on the test sets. While there is some slight increase in PPL during the catastrophic forgetting test, as seen in the results, this impact is minimal compared to the substantial gains in learning the new language distributions. This indicates that the model retains its generalization ability even after adapting to new data. 

\subsubsection*{Explore on the Sequence Modeling Structure}
In our current experiments, the modeling function $f$ used in the TTT layers is primarily instantiated as either a MLP or a linear model. During the adaptation process to new language distributions, such as French and Chinese, we observed that across various model sizes, MLP consistently outperformed the linear model in learning new distribution patterns. This suggests that a more expressive $f$ is better equipped to capture the complexities of evolving new distributions.

Future work could focus on developing more sophisticated instantiations of $f$, particularly in the context of models that integrate LTM. As context length increases, such as in tasks across millions of tokens, $f$ may need to scale in complexity. One promising direction is to explore hierarchical models, where $f$ could incorporate recurrent layers (e.g., RNN or LSTM) or convolutional neural networks that are specifically designed to accumulate and compress information over extended sequences. This would enable the model to continuously learn and adapt, even in environments with extremely long contexts.

Moreover, another interesting research field is multi-level learning to learn. If $f$ itself is a self-attention layer, it could act as a nested inner loop, where each level refines the understanding of the context in relation to LTM data. In this framework, each successive layer of learning could address more abstract or temporally distant features, effectively creating a multi-stage adaptation process that optimizes both short-term and long-term dependencies. This structure would be especially relevant for models aiming to leverage LTM in dynamic or changing environments.

Except for the study of TTT, the evolution towards new model structures aligns well with the goals of model self-evolution and the effective utilization of LTM. By incorporating advanced RNNs with expressive hidden states, there are also some other efficient architectures like Mamba~\cite{He-2024-CrackMamba}, RWKV~\cite{Peng-2024-RWKV}, and biologically-inspired neuron diversity~\cite{Fan-2023-NeuroAI,Network-Model-2024,Training-SNN-2023}, so we can build more sophisticated and efficient models capable of better simulating human-like cognitive processes.

\subsection{Development of LTM-Based Multi-Agent Framework}
\label{subsec:omne}
In Section\ref{chap6:LTM_as_mix}, we employed a hybrid strategy of RAG and training to utilize the generated LTM data, achieving promising performances. Here, we will introduce a LTM-Based Multi-Agent Framework called Omne. Omne is a deeply customized development framework based on the AutoGen Multi-Agent Framework, specifically designed to address the practical application challenges of LTM in AI systems. It extends a range of memory-related infrastructure, including a unified memory model, a multimodal message processing system, and flexible memory storage and manipulation mechanisms. The core goal of Omne is to provide a comprehensive solution that enables the effective deployment of LTM in real-world engineering projects, thereby enhancing the long-term memory capabilities and task-handling efficiency of AI systems.

\subsubsection{Core Modules of Omne} 
\begin{figure}[H] 
\centering \includegraphics[width=0.8\linewidth]{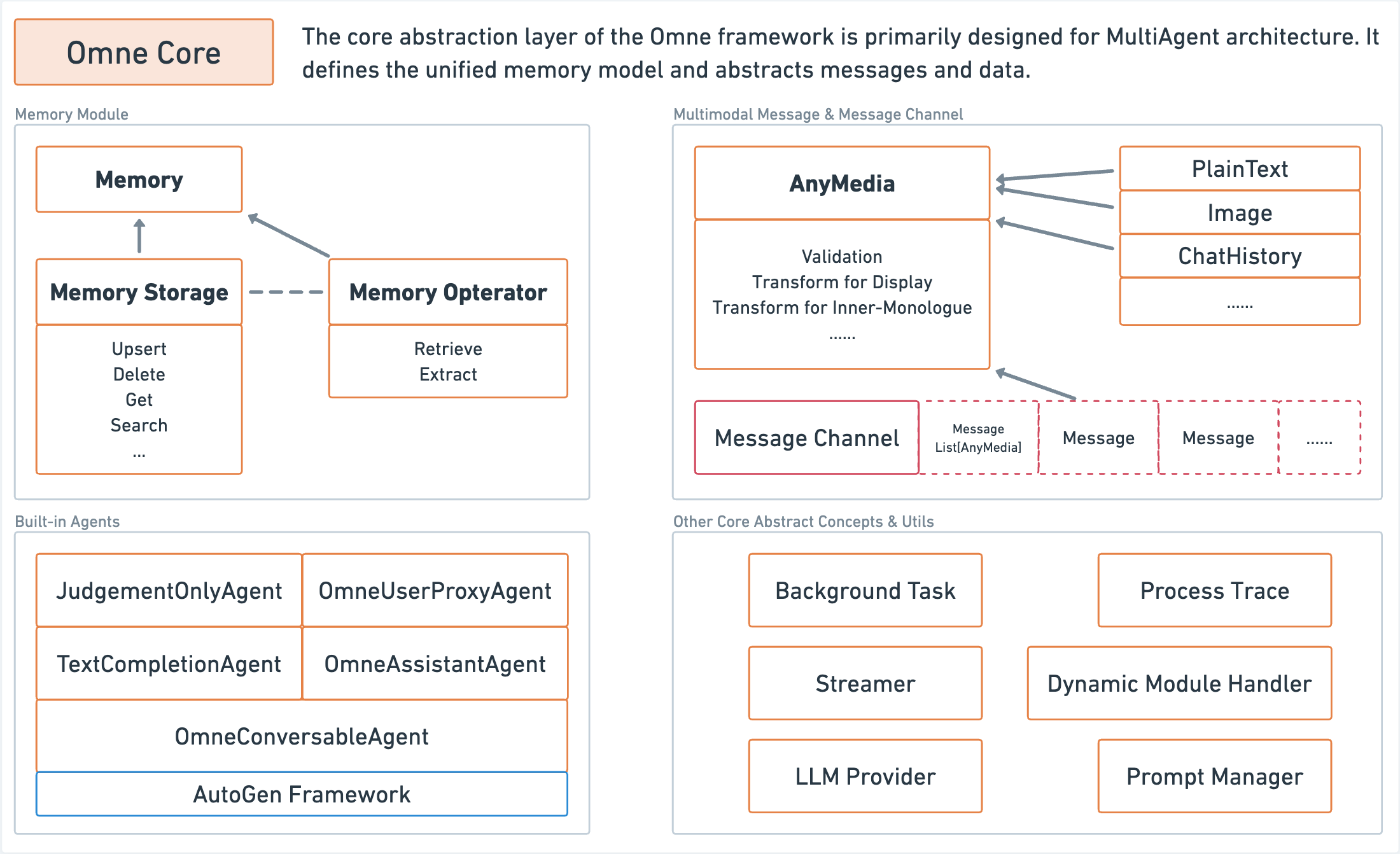} \caption{Illustration of Omne Core, which constitutes of a memory module, multimedia messaging channel, built-in agents and other related tasks.} 
\label{fig_core} 
\end{figure}
Omne Core serves as the primary abstraction layer for the framework, focusing on the design of a multi-agent architecture. It defines a unified memory model and abstracts the processing of messages and data. By offering a standardized memory abstraction mechanism, the Omne framework allows developers to manage memory with consistent operations at any level of the framework.

Coupled with the message channel mechanism provided by the framework layer, Omne enables more complex memory operations, such as asynchronous memory retrieval and on-demand memory extraction and reprocessing across different business scenarios. This design allows Omne to efficiently handle various complex memory and business integration scenarios, offering a robust infrastructure for developing highly intelligent AI applications.

Additionally, whether using traditional Retrieval-Augmented Generation (RAG) techniques or in-context learning approaches, the memory abstraction provided by the Omne framework can be leveraged. This flexibility allows developers to mix and match different technologies to meet specific business requirements and application scenarios.

In terms of multimodal message processing, Omne offers a complete set of message storage and manipulation abstractions. Developers can utilize built-in media types like plain text and images, or extend them based on specific business needs. The framework also supports flexible message visibility definitions to accommodate different business contexts.

For example, when handling chat history that includes both text and images, if using an LLM provider that supports image processing (e.g., GPT-4), the message channel will automatically preserve the images and convert them into a compatible format. If switching to a model that does not support image input (e.g., Llama 3.1), the system will automatically convert images into descriptive text. This design ensures data continuity and usability when transitioning between models with varying capabilities.

\subsubsection{Omne Assistant} 
\begin{figure} [H]
\centering \includegraphics[width=1\linewidth]{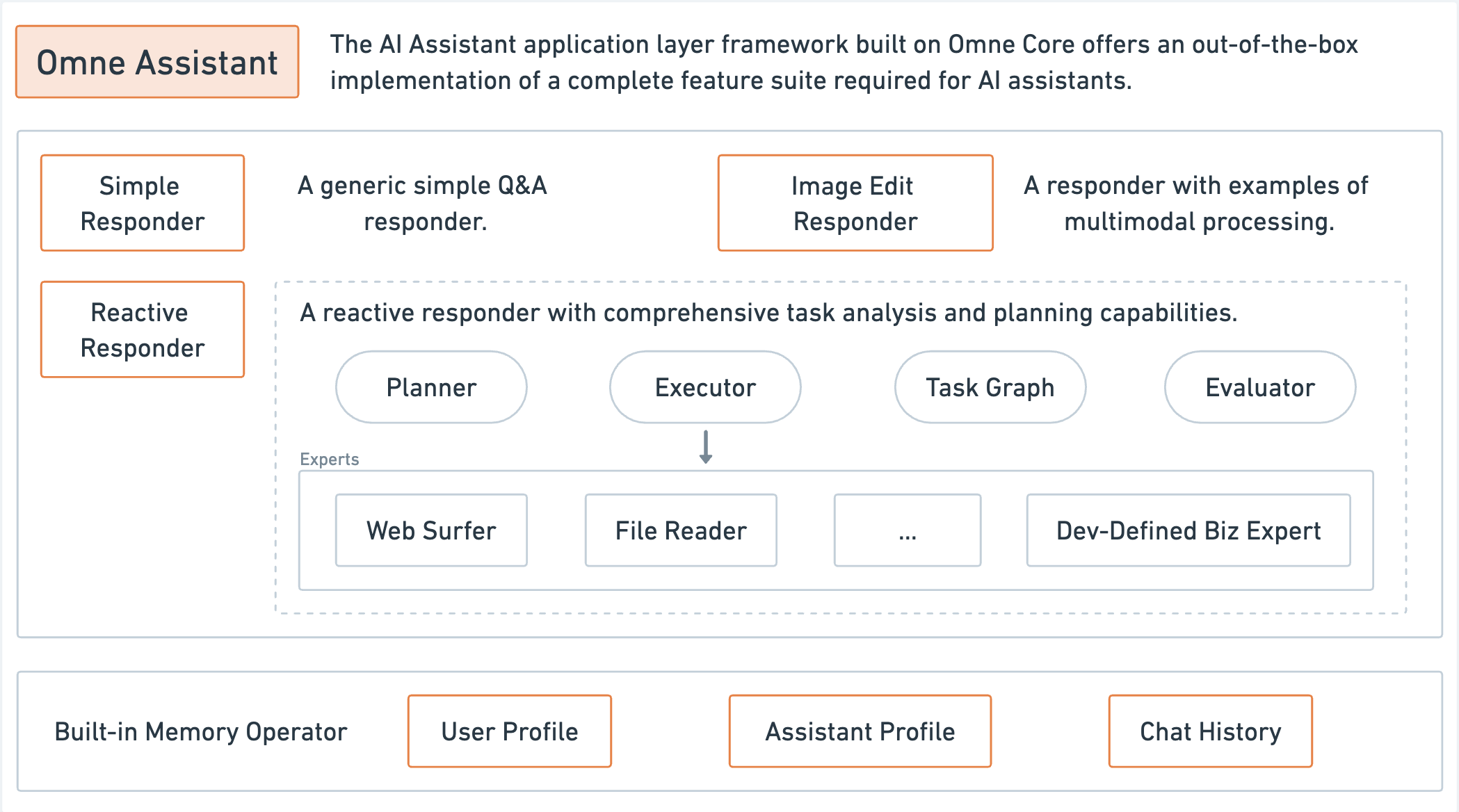} 
\caption{Omne Assistant} 
\label{fig_assis
} 
\end{figure}

Built on Omne Core, Omne Assistant provides a ready-to-use application layer framework specifically designed for the development of AI assistants in chat scenarios. It includes the essential functionalities required for AI assistants, enabling developers to quickly build fully functional chatbots without the need to design foundational components from scratch.

Omne Assistant comes with a \textbf{Simple Responder}, a general-purpose Q\&A responder that can handle basic user chat interactions for instant communication. Additionally, the framework provides a \textbf{Reactive Responder}, which has advanced task analysis and planning capabilities, allowing it to manage more complex user requests that require multi-step reasoning and task orchestration.

With these built-in components, Omne Assistant allows developers to focus on implementing custom functionalities, speeding up the development and deployment of AI assistant applications equipped with long-term memory capabilities.

\subsubsection{Case Study}

In practical applications for planning complex tasks, it is often necessary to balance task quality with generation speed. Therefore, we explore the possibility of allowing sufficiently powerful models to "pre-plan" various tasks and store the results as "memories." During actual online inference, similar tasks can be "recalled," and a faster model can use these pre-planned results to carry out the real task planning.

Using the tools provided by the Omne framework, we can quickly implement In-context Learning (ICL) capabilities in business applications. Below is a simple process outline.
\textit{Usage Example: Enhancing Planning Effectiveness for Highly Complex Tasks Using In-context Learning.}

\begin{itemize} 
    \item Prepare a set of task objectives and pre-constructed task contexts, generating different plans by selecting various models and prompting strategies.[Offline Phase]
    \item Automatically filter effective experiences through simulation of the execution of the planning results.[Offline Phase]
    \item Involve human intervention to assess the validity of the experiences and make adjustments to the results if necessary.[Offline Phase][Optional] 
    \item Save the effective experiences as memories (the simplest method is to combine task objectives and contexts into task descriptions, using these descriptions as vector retrieval indices).[Offline Phase]
    \item Identify user objectives and construct the context needed for task planning.[Online Phase] 
    \item Retrieve "experience memories" to find similar planning experiences and reference these experiences for actual task planning.[Online Phase]
    \end{itemize}

At this point, we have implemented In-context Learning for complex task planning using the Memory Operator and other related suites provided by Omne. It is noteworthy that due to the well-designed framework, the entire process is iterative and minimally invasive to business operations.

The Omne framework is designed and implemented with the goal of applying "memory in a Multi-Agent architecture." It consistently treats "memory" as a primary element and core concept, aiming to build a technical framework that can grow alongside long-term memory technologies and facilitate engineering deployment.

\subsubsection{GAIA benchmark}

We evaluated the Omne framework on the GAIA benchmark \cite{mialon2023gaia}, a rigorous test for general AI assistants comprising over 400 question-answering tasks. These tasks involve complex logical deduction and reasoning processes, requiring capabilities such as numerical calculations, web browsing, video and speech processing, and file manipulation.

To explore the boundaries of AI models, we utilized GPT-4o and o1-preview\cite{openai2024o1preview} within Omne. Omne was equipped with four tools: web surfer, bing search engine, file reader based on llamaparse\cite{llamaindex2024parse} (for various file manipulation), and a logic expert built with o1-preview. Additionally, we implemented a scene router based on o1-preview to assess whether a question could be resolved locally, without external web information and resources. If one question can be solved locally, then we used logic expert to solve it directly(and file reader if needed). Otherwise we turn to a gpt-4o based multi-agent system with all 4 tools mentioned above. Our final results were submitted to official evaluation server  \href{https://huggingface.co/spaces/gaia-benchmark/leaderboard}{official evaluation server} of GAIA in huggingface.

Leveraging OpenAI's GPT-4o and o1-preview as base models, Omne achieved first place (40.53\%) on the test set and second place (46.06\%) on the validation test, establishing a new state-of-the-art on this milestone dataset in AI research. Notably, Omne attained 26.53\% accuracy on the most sophisticated and demanding level 3 questions, demonstrating its potential to solve real-world problems by leveraging powerful foundational models, particularly those with strong reasoning and logical capabilities.

\begin{figure} [H]
\centering \includegraphics[width=1\linewidth]{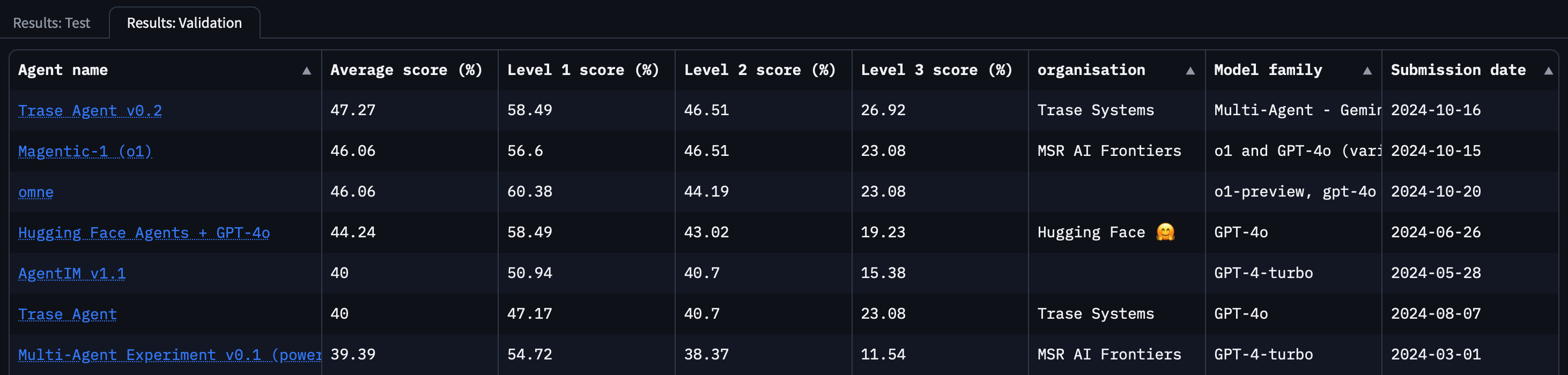} 
\caption{GAIA leaderboard in validation set} 
\label{fig_valid
} 

\centering \includegraphics[width=1\linewidth]{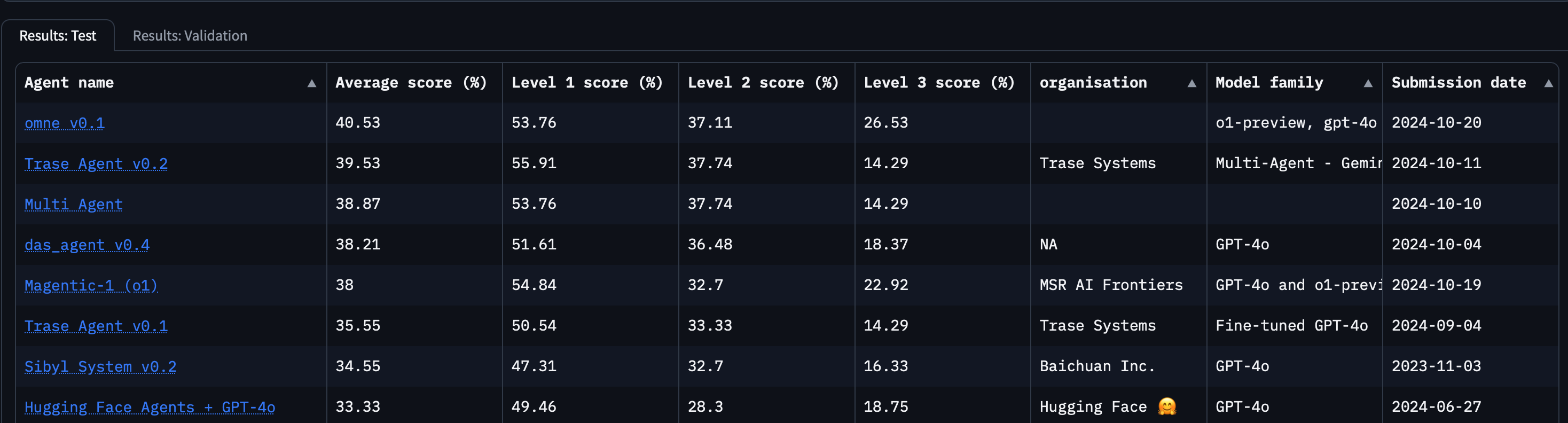} 
\caption{GAIA leaderboard in test set} 
\label{gaia_test
} 
\end{figure}

\section{             Our Future Plans}
\label{chap8}
\label{subsec:main_feature_plan}









As discussed in Section \ref{subsec:main_why_mp}, we believe that model personalization could be one of the key pathways to a second emergence of intelligence. Here, we will briefly describe the key research directions and areas we will focus on in the future, and we welcome more researchers to join us in our efforts to use LTM to improve model personalization. By accumulating data, such as LTM data, different LLM agents could develop diverse and differentiated abilities. The combination of diverse and differentiated individuals is more likely to spark new intelligence. We are also confident that LTM and model personalization could play a pivotal role in achieving this ambitious goal. In the future, we aim to answer the following challenging questions: 
\begin{enumerate}
    \item How to better construct LTM data?
    \item How to Design New Model Architectures for LTM? 
    \item How Can LTM Help Users Ask Better Questions? 
    \item How to Integrate LTM with Inference-Time Search?
    \item How to Use LTM for Agent Self-Evolution in Complex Scenarios?
    \item How to Use LTM in Multi-Agent Scenarios?
\end{enumerate}

\subsection{How to better construct LTM data?}

To better collect longitudinal data, it is essential to establish a system capable of tracking individual data over time. This can be achieved by incentivizing users to continuously engage with data platforms while ensuring privacy and security, encouraging long-term data collection. Additionally, the use of wearable devices and IoT sensors can provide continuous multimodal data collection, which is crucial for building long-term memory models. To balance diversity and data consistency, it is important to curate datasets that represent a wide range of user characteristics while maintaining uniformity in data labeling and collection methods.

We are also advancing our efforts in data synthesis and developing an end-to-end data synthesis system. This system incorporates a continuous learning and adaptive data synthesis framework, allowing it to evaluate the performance of synthetic data and adjust the generation process accordingly. The system includes a feedback loop, enabling it to detect data deficiencies in model personalization and automatically optimize and refine data synthesis methods.

\subsection{How to Design New Model Architectures for LTM?}
One perspective is that as models progress, the increasing length of the context window may provide a pathway toward achieving long-term memory. However, given that the context window is currently limited to the KV cache of short-term sessions, it cannot genuinely achieve cross-task, cross-session long-term memory. We believe that it is necessary to redesign the underlying model architecture, transitioning LLMs from merely relying on "context windows" to a deeper, more structured long-term memory mechanism. The key challenge in this direction is how to enable the model not only to store and retrieve historical information but also to dynamically manage this memory through adaptive adjustments.

Our vision is to conceptualize LLMs as world models, drawing inspiration from the concept of cortical columns, where each layer of the model can be seen as an independent entity. In this way, each layer of the model can dynamically adjust its weights based on new data distributions. This could significantly enhance computational efficiency and provide the model with more flexibility, enabling it to handle more complex and diverse data distributions. Similar to neurons in the brain, different model layers can activate distinct memory fragments based on specific tasks, forming an adaptive reasoning mechanism. We have made preliminary attempts on this idea with TTT, and future research will explore this direction in more depth.

More importantly, the concept of allowing models to autonomously adjust long-term memory based on environmental sensing opens up new avenues for the development of more intelligent AI. Unlike traditional pre-set memory management, models under this paradigm can perceive changes in external environments in real-time and adjust their memory structure accordingly. This adjustment goes beyond simple recall and storage of data and includes the perception and optimization of data distribution changes, ensuring that the model can efficiently and personally adapt to various tasks.

This form of learning not only enhances the model’s adaptability but also enables it to provide more personalized reasoning when facing diverse data. This flexible memory management and optimization mechanism will empower future LLMs to not only better understand user needs but also handle complex reasoning tasks more efficiently. The core of this vision is for the model to optimize its memory management as the environment changes, gradually evolving into an intelligent agent capable of self-learning and optimization. This agent will demonstrate flexibility in handling dynamic and uncertain environments. The integration of long-term memory and model personalization will pave the way for advancements in future AI development.

\subsection{How Can LTM Help Users Ask Better Questions?}

We believe that while LLMs are already capable of providing good responses to user questions, users may struggle to formulate optimal questions, thereby limiting the effectiveness and potential applications of LLMs. 

For example, in a news reading context where a model is expected to generate in-depth questions that stimulate critical thinking and enhance readers' engagement. However, obtaining training data that enables a model to ask "high-quality" questions poses a significant challenge. One potential approach to generating such data is to extract and summarize questions from news interview programs. These questions are typically crafted by experienced hosts and journalists, who are adept at posing thought-provoking and insightful queries. By analyzing and summarizing these questions alongside the relevant news content, we can construct a dataset that embodies the key characteristics of high-quality inquiry, such as probing assumptions, exploring diverse perspectives, and addressing complex issues.

Exploration is a key method by which agents discover new possibilities, while pruning accelerates convergence and avoids ineffective paths. Striking a balance between exploration and pruning is crucial to improving an agent’s ability to ask better questions.

How can an agent find effective strategies during initial exploration and then quickly prune ineffective paths in the later stages? Could we introduce a dynamic exploration-pruning strategy, where the agent leans more toward exploration during the early stages of high uncertainty and accelerates pruning as certainty increases? When facing new problems, how can an agent leverage long-term memory to transfer past strategies from similar problems to the current task, thus reducing exploration time and resource consumption?

How can long-term memory guide pruning? Can agents identify inefficient or unfeasible paths based on past exploration results, allowing them to skip these paths in new tasks? For instance, in a medical system, an agent could quickly prune ineffective treatment plans based on past data, focusing instead on more effective strategies. How can we balance local exploration with global pruning? In multi-agent systems, one agent's exploration may influence the strategies and outcomes of others. How can we ensure that the pruning process does not hinder other agents’ exploration or the evolution of the overall system?

\subsection{How to Integrate LTM with Inference-Time Search?}

A major advance in OpenAI's O1 model is the integration of reinforcement learning (RL) to enhance LLMs' multi-step reasoning and planning capabilities. Specifically, RL introduces a powerful inference-time search mechanism, enabling the pretrained model to effectively "think" during the inference phase. By generating multiple reasoning trajectories and selecting the optimal one, the model can refine its reasoning process in real-time. This combination of "memorization" (long-term memory, LTM) and "thinking" (dynamic reasoning) will jointly define the future intelligence level of AI systems.

We pose the following questions for further exploration:

How can the dynamic evolution of long-term memory (LTM) enhance LLM's reasoning and search abilities?
The dynamic evolution of LTM could play a critical role in supporting more robust and adaptive reasoning. LTM allows a model to not only store but also update knowledge across interactions, which could directly improve reasoning by providing more relevant historical context and insights. LTM could act as a repository for key information, reducing the need for the model to "relearn" common knowledge with each inference. This could allow the LLM to avoid repeating mistakes or inefficient paths it had previously explored. By evolving this memory over time, the LLM can avoid redundant or inefficient search paths, guiding it toward more relevant reasoning trajectories.

Can LTM be useful for representing LLMs' thought processes as a sequence of state abstractions, so that reinforcement learning can efficiently explore the large space of reasoning paths?
We believe LTM could provide as a structured representation of the LLM's thought processes, capturing key intermediate states during multi-step reasoning. By encoding these states into meaningful abstractions, LTM can create a more tractable search space for RL to navigate. This would allow the reinforcement learning agent to focus on important or novel reasoning paths, making exploration more efficient and targeted. Essentially, LTM could help map the vast reasoning space into manageable, hierarchical abstractions that RL can use to prune less optimal trajectories and focus on those with higher potential outcomes.

How can reinforcement learning-based search methods enhance LTM?
RL can improve LTM by augmenting it through real-time exploration uring the inference phase. For example, if the model's memory lacks certain knowledge or if its reasoning process encounters ambiguities, an RL-driven search method could generate targeted questions or reasoning trajectories to actively fill the gaps in memory. 

Consider a scenario where an AI system encounters a new task or topic that challenges its current knowledge base. The RL-based approach could iteratively search for relevant questions that would retrieve or generate knowledge to augment LTM. For another example, RL could evaluate different strategies for updating LTM by comparing how changes in memory influence the success of reasoning over time. By using feedback signals from multiple reasoning trajectories, the RL agent can continuously refine how the memory is updated and leveraged, ensuring that the memory evolves in a way that enhances future reasoning tasks.

\subsection{How to Use LTM for Agent Self-Evolution in Complex Scenarios?}

In the previous discussions, we mentioned that LTM plays a crucial role in the capability evolution of agents, particularly with the additional LTM gained from interaction history and reflective reasoning, which can directly enhance task performance. However, these improvements come with a prerequisite: we must have ground-truth or a well-established evaluation mechanism to provide feedback, guiding the accumulation of the agent's LTM.

Nevertheless, in complex task scenarios, on one hand, collecting a large amount of high-quality ground-truth is very challenging, often requiring high-quality annotations from experts and significant participation. On the other hand, building evaluators for individual tasks is also very difficult, especially in generative tasks, where effective evaluation methods are currently lacking. Therefore, how to use LTM for agent self-evolution in complex systems remains a vital challenge.

We believe a feasible research approach is to achieve self-evolution through environmental feedback, thereby enabling effective accumulation and use of LTM. Specifically, environmental feedback may come from the physical world, the Simulacrum world, or a combination of both. The evolution in the physical world is closely related to embodied AI research, but the evolution efficiency is relatively low; constructing a Simulacrum world is very challenging, but it offers higher efficiency and lower cost. Moving forward, we will conduct in-depth exploration on Agent Self-Evolution combined with LTM.

\subsection{How to Use LTM in Multi-Agent Scenarios?}
We believe that multi-agent cooperation may lead to the second time of intelligence emergence, which is the ultimate goal we pursue for agent self-evolution. In multi-agent systems, individual agents' actions are typically discrete and local, but the system requires them to co-evolve to achieve global optimization. Agents must use long-term memory for task planning while continuously exploring and pruning to enable overall evolution. However, achieving this goal presents a series of intermediate challenges, especially in the accumulation and utilization of LTM during multi-agent collaboration. Specifically, the following points are crucial.

First, how can we accumulate individual LTMs for collaborating agents? In collaborative scenarios, since multiple agents are responsible for different parts of a complex task, the final completion of the task cannot be simply used as direct feedback for the accumulation of each agent's LTM. Instead, an additional evaluation and decomposition strategy is required to convert it into feedback for each stage, similar to a step-grained reward rather than an overall reward (especially when this reward may be binary). This is critical for the co-evolution of multiple agents.

Second, should there be a certain mechanism for sharing and utilizing LTM during multi-agent collaboration? Currently, there are two forms of multi-agent collaboration: one where each agent completes a sub-module of the task, and another where multiple agents collaborate through discussion to make the final decision. Regardless of the specific form of cooperation, if we can introduce some form of shared memory or communication mechanism to help agents understand the decisions and outcomes of other agents, allowing them to consider global objectives more effectively when making local decisions, we believe it will directly and effectively contribute to enhancing overall capabilities.

We believe that addressing the above issues will significantly advance the evolution of multi-agent collaboration.

\section{Conclusion}
\label{subsec:main_conclusion}



In this report, we propose that long-term memory mechanisms and the evolving nature of knowledge will be crucial. Current models treat all data uniformly, from ancient to modern, without capturing the gradual progression of knowledge. Human cognition, however, evolves—children learn from simple to complex concepts, and knowledge builds upon itself over time. This temporal structuring of data, where models learn through a progression of difficulty or sequence, could help them grasp not just static facts but the relationships and evolution of knowledge itself.

By leveraging longer-term memory architectures, models could begin to capture this evolutionary aspect of learning. Instead of simply memorizing information, they could learn the dynamics of knowledge development, allowing them to operate over extended temporal scales. This would also involve recursive learning, where feedback from real-world environments shapes the model’s growth, leading to a self-boosting mechanism. As neural networks evolve, they may focus more on fast pruning of search spaces during gradient descent and benefit from enhanced parallelism in recursive computations. Ultimately, scaling laws will persist, but model architectures will need to adapt to handle increasing complexity while maintaining efficiency.

Looking forward, model personalization, inspired by human diversity, could drive a new level of intelligence. In multi-agent systems, diverse, LTM-powered agents could collaborate more effectively, balancing exploration and pruning to achieve global optimization. This personalization and adaptability may be key to fostering a second emergence of intelligence, where agents evolve together to tackle increasingly complex problems.This exploration into long-term memory and evolutionary learning will drive advancements in AI, particularly for models that need to continuously adapt, learn, and personalize over time. In this paper, we aim to articulate our vision and roadmap for future research directions and critical focus areas. We invite fellow researchers to engage with our findings and collaborate with us in the pursuit of utilizing LTM to enhance model personalization.

\pagebreak




\bibliography{ltm_ref}
\bibliographystyle{unsrt}

\newpage
\appendix
\section{RTG prompt}
\label{appendix_A}

Figure \ref{fig:rtg_prompt} shows the prompt we use to perform RTG. 

\begin{center}
\label{fig:rtg_prompt}
\fcolorbox{black}{gray!10}{\parbox{.9\linewidth}{
\textit{\{contexts\}}

\textit{\{question\}}

Answer the question using the information given in the documents. You should respond in JSON format with four keys: `Relevant': list[int] or `None', `Cited': list[int] or `None', `Grounded': str, and `Answer': str.
Carefully perform the following instructions, in order:

- Firstly, decide which of the documents are relevant to question. You should be thorough in curating the list to ensure all relevant information is included. You should answer with a comma-separated list of document numbers. If none are relevant, you should instead write `None'. Use the JSON key: ``Relevant'.

- Secondly, decide which of the documents contain facts that should be cited in a good answer to the user's question. You should be thorough in curating the list to ensure all relevant information is included. You should answer with a comma-separated list of document numbers. If you dont want to cite any of them, you should instead write `None'. Use the JSON key: `Cited'.

- Thirdly, provide a comprehensive step-by-step reasoning on how to answer the question and cite all the `Cited Documents' in your reasoning. In your reasoning, copy paste the cited text and include them in <begin\_cite: doc\_num> and <end\_cite: doc\_num> when the quotes come from a document. This would mean that things outside of <begin\_cite: doc\_num> and <end\_cite: doc\_num> are not directly copy paste from the documents. e.g. <begin\_cite: 0>my fact<end\_cite: 0> for a quote from Document 0. Use the JSON key: `Grounded'.

- Finally, provide a detailed and complete answer to user's question in high quality natural English. Use your reasoning and cited documents to help you. Do not insert any citations or grounding markup. Use the JSON key: `Answer'}}
\end{center}

\end{document}